\documentclass[runningheads]{llncs}

\usepackage{eccv}

\usepackage{eccvabbrv}

\usepackage{graphicx}
\usepackage{booktabs}
\usepackage{algorithm}
\usepackage{wrapfig}
\usepackage{listings}
\usepackage[noend]{algpseudocode}
\usepackage{capt-of} 
\usepackage{multirow} 
\usepackage{tikz}
\usepackage{pifont}
\usepackage{makecell}

\newcommand{\datasetname}{\textsc{VisReason}}
\newcommand{\datasetnamemax}{\textsc{VisReason-Pro}}
\newcommand{\modelname}{\textbf{VisReason}}

\usepackage[accsupp]{axessibility}  

\usepackage{hyperref}

\usepackage{orcidlink}

\begin{document}

\title{\textsc{VisReason}: A Large-Scale Dataset for Visual Chain-of-Thought Reasoning}

\titlerunning{\textsc{VisReason}}

\author{
Lingxiao Li\inst{2}\orcidlink{0009-0009-2297-5822} \and
Yifan Wang\inst{1}\orcidlink{0009-0005-1567-2787} \and
Xinyan Gao\inst{3}\orcidlink{0009-0009-9297-4331} \and
Chen Tang\inst{3}\orcidlink{0000-0002-0108-6729} \and
Xiangyu Yue\inst{3}\orcidlink{0000-0002-6887-2046} \and
Chenyu You\inst{1}\orcidlink{0000-0001-8365-7822}\thanks{Corresponding author.}
}

\authorrunning{L. Li et al.} \institute{ \textsuperscript{1}Stony Brook University, Stony Brook, NY, USA\\ \quad \textsuperscript{2}Boston University, Boston, MA, USA\\ \quad \textsuperscript{3}MMLab, The Chinese University of Hong Kong, Hong Kong SAR, China\\
\email{lxli@bu.edu, chenyu.you@stonybrook.edu}\\
\small{\url{https://y-research-sbu.github.io/VisReason/}}
}

\maketitle 



\begin{center}
\captionsetup{hypcap=false}
\includegraphics[width=0.94\linewidth]{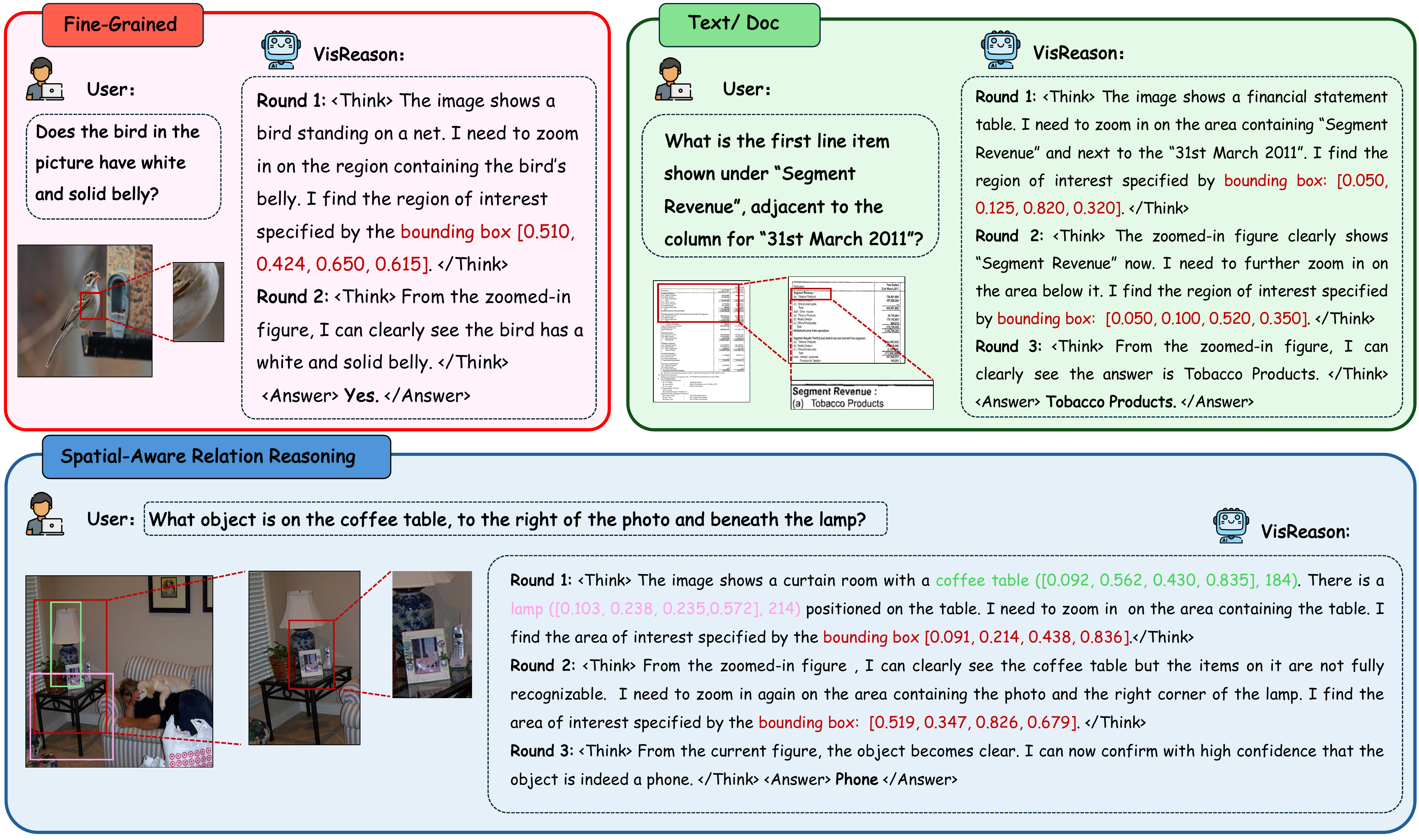}
\captionof{figure}{ An MLLM fine-tuned on {\datasetname} emulates a \textit{human-like visual reasoning process} to solve a complex query. Rather than processing the entire image uniformly, the model adopts a \textbf{dynamic global-to-local workflow}: it first assesses the overall scene, then progressively focuses on salient regions to collect fine-grained visual evidence. This multi-step, spatially grounded \textit{visual Chain-of-Thought} allows the model to anchor its reasoning in localized visual cues, supporting fine-grained recognition and spatial reasoning tasks that are difficult to solve from a single global view.}
\label{fig:illustration_examples}
\end{center}


\begin{abstract}
\label{sec:abstract}

Chain-of-Thought (CoT) prompting has proven remarkably effective for eliciting complex reasoning in large language models (LLMs). Yet, its potential in multimodal large language models (MLLMs) remains largely untapped, hindered by the absence of large-scale datasets that capture the rich, spatially grounded reasoning intrinsic to visual understanding. Existing visual-CoT resources are typically small, domain-specific, or lack the structured stepwise supervision necessary for compositional visual reasoning.
In this paper, we introduce~\datasetname{}, a large-scale dataset designed to advance visual Chain-of-Thought reasoning. VisReason comprises 489K annotated examples spanning four diverse domains, each featuring multi-round, RoI-grounded rationales that guide MLLMs through interpretable visual reasoning steps. Building upon this, we curate \datasetnamemax{}, a 165K subset produced with a stronger GPT annotator, enriched with detailed reasoning traces and depth-augmented spatial annotations derived from monocular depth and segmentation cues. Fine-tuning strong MLLM backbones on VisReason and VisReason-Pro yields substantial improvements in step-by-step visual reasoning accuracy, RoI localization, interpretability, and fine-grained/spatial reasoning performance. These results demonstrate that VisReason equips MLLMs with more systematic and verifiable visual reasoning capabilities. We envision VisReason as a cornerstone for cultivating human-like visual reasoning, paving the way toward the next generation of multimodal intelligence.

\keywords{Multimodal Large Language Models \and Visual Chain-of-Thought Reasoning \and Spatially Grounded Reasoning}

\end{abstract}
\section{Introduction}
\label{sec:intro}
Recent multimodal large language models (MLLMs) have achieved remarkable progress in practical visual understanding, largely by pairing high-capacity language models with powerful visual encoders through sophisticated alignment pipelines~\cite{2023GPT4VisionSC,zhu2023minigpt,yin2023survey,bai2023qwen}. Foundational models such as LLaVA~\cite{li2024llava}, InternVL~\cite{chen2024internvl}, Qwen2.5-VL~\cite{bai2025qwen25vl}, and MiniCPM-V~\cite{yao2024minicpm}, have demonstrated state-of-the-art performance across a wide range of tasks, including visual question answering~\cite{li2024llava}, fine-grained grounding~\cite{peng2024kosmos2}, and optical character recognition~\cite{zhang2023llavar}. These systems have rapidly evolved into versatile visual reasoning assistants for knowledge access, creative work, and embodied interaction.

However, despite architectural advances, the reasoning paradigm of most MLLMs remains rudimentary. In text-only LLMs, Chain-of-Thought (CoT) prompting has revolutionized complex reasoning by training models to articulate explicit intermediate rationales before producing an answer~\cite{wei2022chain}. This stepwise supervision has yielded dramatic improvements in arithmetic, commonsense, and symbolic reasoning. In contrast, the multi-modal domain has yet to experience an equivalent shift~\cite{liu2024chain,shao2024visualcot,man2025argus}. Current MLLMs are predominantly optimized in a direct input-to-answer manner, offering no guidance on intermediate cognitive steps. 
Although recent work has begun exploring multi-round visual CoTs~\cite{qi2024cogcom,su2025pixel}, closely aligned with the broader ``thinking-with-images'' paradigm~\cite{openai2024thinkingwithimages,su2025thinking}, these attempts still lack the scale, domain breadth, and spatial grounding necessary to model global-to-local visual reasoning. These deficiencies encourage shortcut learning, over-reliance on linguistic priors, and vulnerability to hallucination when confronted with multi-step visual queries~\cite{ke2025explain}.


We argue that this limitation arises from a \textit{fundamental misalignment} between prevailing training paradigms and the structure of human visual cognition. Humans solve visually grounded problems through a global-to-local reasoning process: first forming a holistic hypothesis about the scene, then iteratively narrowing focus to inspect salient regions and refine local evidence. This involves targeted manipulations such as zooming or cropping to probe inter-object relations and depth cues until the task is resolved~\cite{qi2024cogcom}. While modern VLMs have acquired basic perceptual skills like grounding and OCR during pretraining, they lack supervision that captures this human-like reasoning workflow. 
The root cause is the inadequacy of current visual reasoning datasets, which suffer from three key deficiencies:
\ding{182} limited scale and domain diversity, constraining the ability to learn generalizable reasoning patterns \cite{zhang2025cof,man2025argus,sarch2025grounded};
\ding{183} insufficient multi-round supervision, often collapsing reasoning into a single-step QA pair without explicit rationales \cite{li2025vocot,ye2024voco,qi2024cogcom}; and 
\ding{184} predominantly 2D annotations that provide limited supervision for relative-depth and occlusion-aware spatial reasoning~\cite{shao2024visualcot,sarch2025grounded,wu2025gcot}.
This lack of high-quality, process-level data has become a central obstacle to developing MLLMs capable of explicit, grounded visual reasoning rather than shortcut-based perception.

To address these challenges, we introduce \datasetname{}, a large-scale dataset explicitly designed to instill human-like, spatially grounded reasoning in MLLMs. \datasetname{} consists of 489K annotated examples across four domains, complemented by {\datasetnamemax{}}, a 165K-example expert-curated subset with richer rationales and depth-informed supervision. To build these, our unified construction pipeline begins by enriching each image with pseudo-depth cues derived from monocular depth estimation and semantic segmentation. Using these depth-augmented scene representations, we prompt advanced MLLMs to generate multi-round visual CoT traces that follow global-to-local reasoning workflows. For \datasetnamemax{}, we employ stronger GPT-4.1-series guidance to produce detailed rationales, ensuring higher semantic fidelity and reasoning quality. Each step provides a concise scene summary, identifies a region of interest, and concludes with an explanatory rationale. This fine-grained supervision discourages shortcut learning, promotes global-to-local ``zoom-and-verify'' behavior, and improves fine-grained and spatial reasoning. Furthermore, we incorporate ordinal-depth cues -- most prominently in \datasetnamemax{} -- to encourage spatial reasoning from single-view depth estimates without claiming metric 3D reconstruction. Building upon this resource, we establish a comprehensive evaluation suite to assess models' stepwise reasoning, RoI localization, and depth-augmented spatial reasoning abilities, offering a unified platform for advancing visual CoT research.

Our main contributions are as follows:
\begin{itemize}
\item We construct and release \datasetname{}, a large-scale dataset of 489K examples across four diverse domains, together with its 165K high-quality subset {\datasetnamemax{}} produced under stronger GPT-4.1-series guidance. The dataset provides multi-round, RoI-grounded step-by-step supervision enriched with depth-informed annotations for relative-depth and spatial reasoning.
\item We establish a corresponding evaluation suite based on \datasetnamemax{}, designed to evaluate models' fine-grained reasoning, RoI localization, and depth-augmented spatial reasoning through detailed, multi-step visual CoT tasks.
\item We design a unified training and inference pipeline that leverages these annotations to instill spatially grounded CoT reasoning in MLLMs. Experiments across strong MLLM backbones demonstrate improved fine-grained recognition, spatial relation reasoning, RoI grounding, interpretability, and transfer beyond a single base model.
\end{itemize}

\section{Related Works}
\subsection{Multimodal Large Language Models}
Multimodal large language models (MLLMs) have become a central direction in vision-language research~\cite{liu2023visual,sun2026coma,you2020contextualized,you2021mrd,you2024calibrating,liu2021aligning,chen2021self,liu2023medical,liu2025together,you2025uncovering}. Most modern systems couple a high-capacity visual encoder, such as ViT~\cite{dosovitskiy2020image}, with a projection module, \textit{e.g.}, an MLP or Q-Former~\cite{li2023blip}, to align visual representations with the language embedding space of an LLM for autoregressive decoding~\cite{yang2024qwen2}. Recent model families, including LLaVA-OneVision~\cite{li2024llava}, InternVL~\cite{chen2024internvl}, Qwen-VL~\cite{bai2023qwen}, LLaVA-UHD~\cite{xu2024llava}, InternVL-3~\cite{zhu2025internvl3}, Qwen2.5-VL~\cite{bai2025qwen25vl}, GPT-4~\cite{2023GPT4VisionSC}, and Gemini-2.5-Pro~\cite{gemini2025report}, have substantially advanced multimodal reasoning and high-resolution perception.
Despite this progress, current MLLMs still struggle when answer-critical evidence is small, visually subtle, or embedded in complex spatial relations. A uniform high-resolution encoding pipeline often allocates computation to irrelevant regions and may amplify linguistic priors rather than support deliberate visual inspection. This motivates adaptive reasoning workflows that decide where to look, when to zoom, and how to verify local evidence, which is the focus of this work.

\subsection{Multimodal Reasoning}
Chain-of-Thought (CoT) prompting has improved the interpretability and reliability of LLM reasoning~\cite{wei2022chain,zhang2022automatic,zhang2025tokenization,gao2025mat}, inspiring analogous efforts in MLLMs. Existing methods can be broadly grouped into model-oriented and dataset-oriented approaches. Model-oriented methods elicit visual reasoning through in-context, tool-based, or compositional prompting~\cite{zhang2024prompt,mitra2024compositional,sun2024coma,gupta2023visual,gao2024clova,chen2024visual}, or improve reasoning behavior through supervised fine-tuning and reinforcement learning~\cite{liu2025visual,sun2025ouroboros,yang2025r1,zhou2025r1,zhang2025r1,liu2025othink}. These approaches encourage intermediate rationales, but typically provide limited explicit supervision over stepwise visual attention.

Dataset-oriented efforts instead supervise the reasoning process more directly. However, many prior datasets reduce visual reasoning to a single localization or attention step, such as $V^{\star}$, VPD, Visual CoT, DualFocus, and Chain-of-Spot~\cite{wu2024vstar,hu2024visual,shao2024visualcot,cao2024dualfocus,liu2024chain}. This simplified supervision can leave multi-step reasoning under-specified and encourages shortcuts when the task requires iterative evidence gathering. More recent work explores multi-round, manipulation-based, or interleaved visual reasoning, including CogCoM and VoCoT~\cite{qi2024cogcom,li2025vocot}, visually grounded region replay in VGR~\cite{wang2025vgr}, interleaved text-image traces in Zebra-CoT~\cite{li2025zebracot}, and code-driven visual thoughts in CodePlot-CoT~\cite{duan2025codeplot}. Related efforts further study adaptive visual search, grounded long-chain reasoning, and pixel-space reasoning~\cite{zhang2025cof,dong2025insightv,man2025argus,su2025pixel}, aligning with the broader ``thinking with images'' paradigm.

Nevertheless, existing dataset-oriented resources still face a trade-off among scale, domain coverage, explicit multi-round RoI supervision, and spatial grounding. In particular, most visual-CoT datasets remain primarily 2D and provide limited supervision for ordinal-depth or occlusion-aware spatial relations~\cite{shao2024visualcot, li2024bifrost, sarch2025grounded,wu2025gcot}. In contrast, VisReason provides a large-scale, multi-domain benchmark with compact global-to-local reasoning traces, RoI-grounded supervision, and depth-informed annotations, enabling MLLMs to learn more faithful fine-grained and spatial reasoning workflows.

\begin{figure*}[t]
\centering
\includegraphics[width=0.88\linewidth]{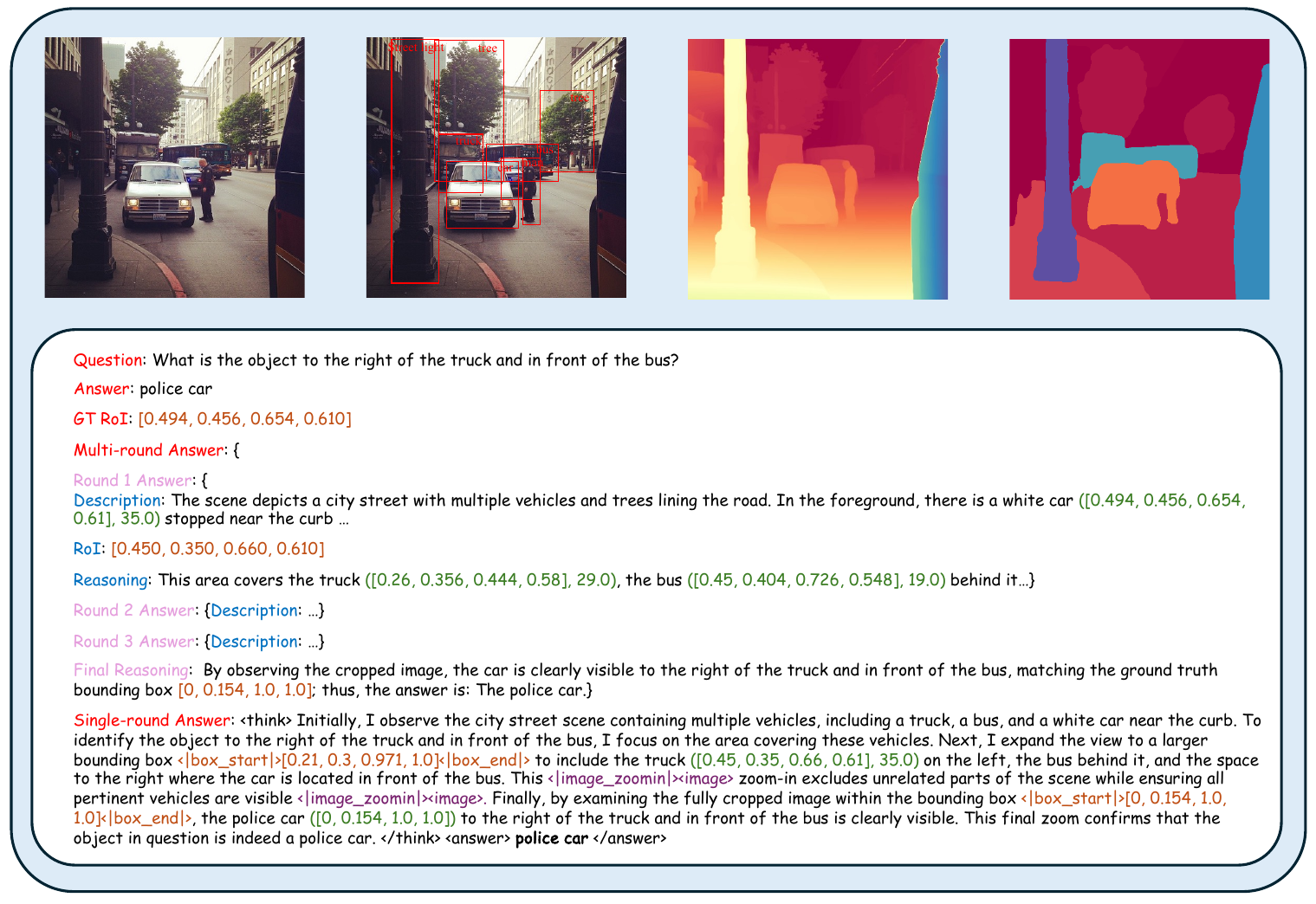}
\caption{For each image-question pair, we provide a region of interest (bounding box) and a compact multi-round visual chain-of-thought: each round offers a scene sketch, an optional zoom to a predicted RoI, and a brief rationale. When available, depth cues indicate ordinal ordering. The annotations are concise and process-oriented, enabling spatially grounded reasoning on fine details and complex relations.}
\label{fig:data_demo}
\end{figure*}

\section{VisReason}
As detailed in \cref{sec:intro}, existing visual reasoning datasets suffer from three persistent limitations -- insufficient scale and domain coverage, lack of multi-round stepwise supervision, and limited supervision for depth-augmented spatial reasoning -- necessitating a resource that trains MLLMs to follow global-to-local visual reasoning. We address these gaps by curating \textbf{\datasetname{}} -- a large, spatially aware visual chain-of-thought (CoT) corpus that explicitly supervises the \emph{process} of visual reasoning rather than only final answers. As illustrated in \cref{fig:data_demo}, each sample consists of a question, an answer, and a \emph{multi-round} CoT that mirrors global-to-local problem solving: every round provides (i) a brief \emph{scene description}, (ii) a predicted \emph{region of interest} (RoI, via bounding box) when zoom is warranted, and (iii) a detailed \emph{rationale} explaining why that RoI suffices. Importantly, zoom-in operations are triggered only when the target object or text is small or visually subtle, allowing the model to retain its efficiency and reliability on simpler cases while engaging in fine-grained reasoning when necessary. Beyond 2D cues, we attach pseudo-depth and segmentation signals -- monocular depth and semantic segmentation -- so that the chain can reference ordinal depth and part/region evidence when needed. This unified annotation format encourages models to localize, zoom, and verify iteratively, reducing shortcut learning and promoting depth-augmented spatial reasoning.

To ensure broad coverage while keeping the focus on process supervision, \datasetname{} spans \textit{four domains} -- text/doc understanding, fine-grained recognition, general VQA, and spatial-aware relational reasoning -- continuing and extending prior category choices (\cref{tab:dataset}). In total, the primary set contains \textbf{489k} examples, and we further release a \textbf{165k} high-fidelity subset, \datasetnamemax{}, with richer rationales and stronger depth-informed grounding. Together, these resources offer detailed, stepwise supervision (see \cref{fig:data_demo}) designed to cultivate global-to-local “zoom–and–verify” behaviors and robust reasoning over small objects, 2D spatial relations, and ordinal-depth relations.

\subsection{Dataset Generation}
\noindent\textbf{\datasetname{}.}~~
Building on the Visual-CoT seed~\cite{shao2024visualcot}, we expand each image–question–answer triple with \emph{process-level} supervision. For every example, the model (GPT-4.1-Nano~\cite{achiam2023gpt}) is prompted to produce a concise scene description, a normalized region of interest (RoI; $[x_1, y_1, x_2, y_2] \in [0,1]^4$), and a brief rationale. We enforce coverage by adjusting the RoI to tightly contain the ground-truth box and iteratively refining via global-to-local zoom. The refinement terminates when the RoI area is no more than twice the GT area or when a small round budget ($\leq3$) is reached. For easier cases where the target object occupies a sufficiently large portion of the image (\eg, $>30\%$ of the area), we skip iterative cropping and instead provide a single detailed reasoning step followed directly by the final answer. This yields multi-round chains that are compact yet faithful, providing stepwise evidence aligned with the final answer while discouraging shortcut learning.

\begin{figure*}[t]
\centering
\includegraphics[width=0.95\linewidth]{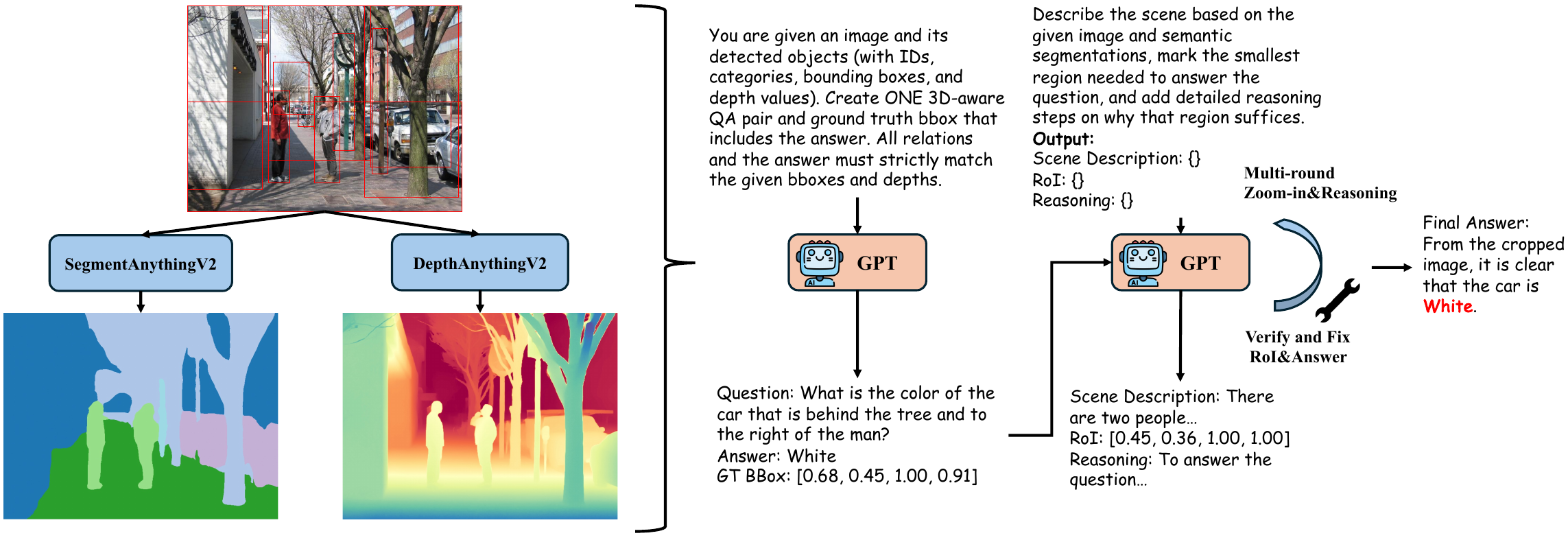}
\caption{Pipeline for \datasetname{} and \datasetnamemax{} data generation and supervision. Given an input image, we derive semantic segments and monocular depth to form an object list with categories, bounding boxes, and ordinal depth; a generator then produces a depth-augmented spatial QA pair and target box. A second stage emits a compact, multi-round visual CoT -- scene sketch, predicted RoI, and rationale -- while iteratively zooming and verifying (with RoI/answer fix) until the final answer and finalized annotations are obtained.}

\label{fig:pipeline}
\end{figure*}

\noindent\textbf{\datasetnamemax{}.}~~
As illustrated in \cref{fig:pipeline}, \datasetnamemax{} augments the above pipeline with explicit spatial priors to elicit depth-augmented reasoning. This subset is constructed primarily from the GQA portion of Visual-CoT, which provides richer annotations (\textit{e.g.}, bounding boxes, relations). We first derive pseudo-depth and segmentation cues per image -- monocular depth and semantic segmentation (object IDs, categories, pixel boxes, and ordinal depths) -- and feed these structured signals, together with the image, to a stronger generator (GPT-4.1-Mini~\cite{achiam2023gpt}). The model is instructed to create depth-augmented questions whose relations jointly involve 2D layout (\emph{left of}, \emph{above}) and ordinal depth (\emph{in front of}, \emph{behind}), outputting a consistent GT box for the target. We then apply the same verify-and-fix routine with multi-round zoom (round $\leq4$) to obtain concise descriptions, RoIs, and rationales at each step. In addition to these \emph{multi-round} traces, we provide a \emph{single-round} distilled variant that compacts the multi-step chain into one rationale and a final RoI. As shown by the \emph{Single-round Answer} in \cref{fig:data_demo}, our \datasetnamemax{} also enables single-pass answering while preserving explicit process supervision. The result is a depth-informed visual CoT corpus tailored for small-object queries, 2D spatial relations, and ordinal-depth reasoning. More details of the prompt design and algorithms are provided in the Supplementary Material.

\subsection{Dataset Analysis}
\label{sec:dataset analysis}
\begin{figure*}[t!]
  \centering
  \includegraphics[width=0.95\linewidth]{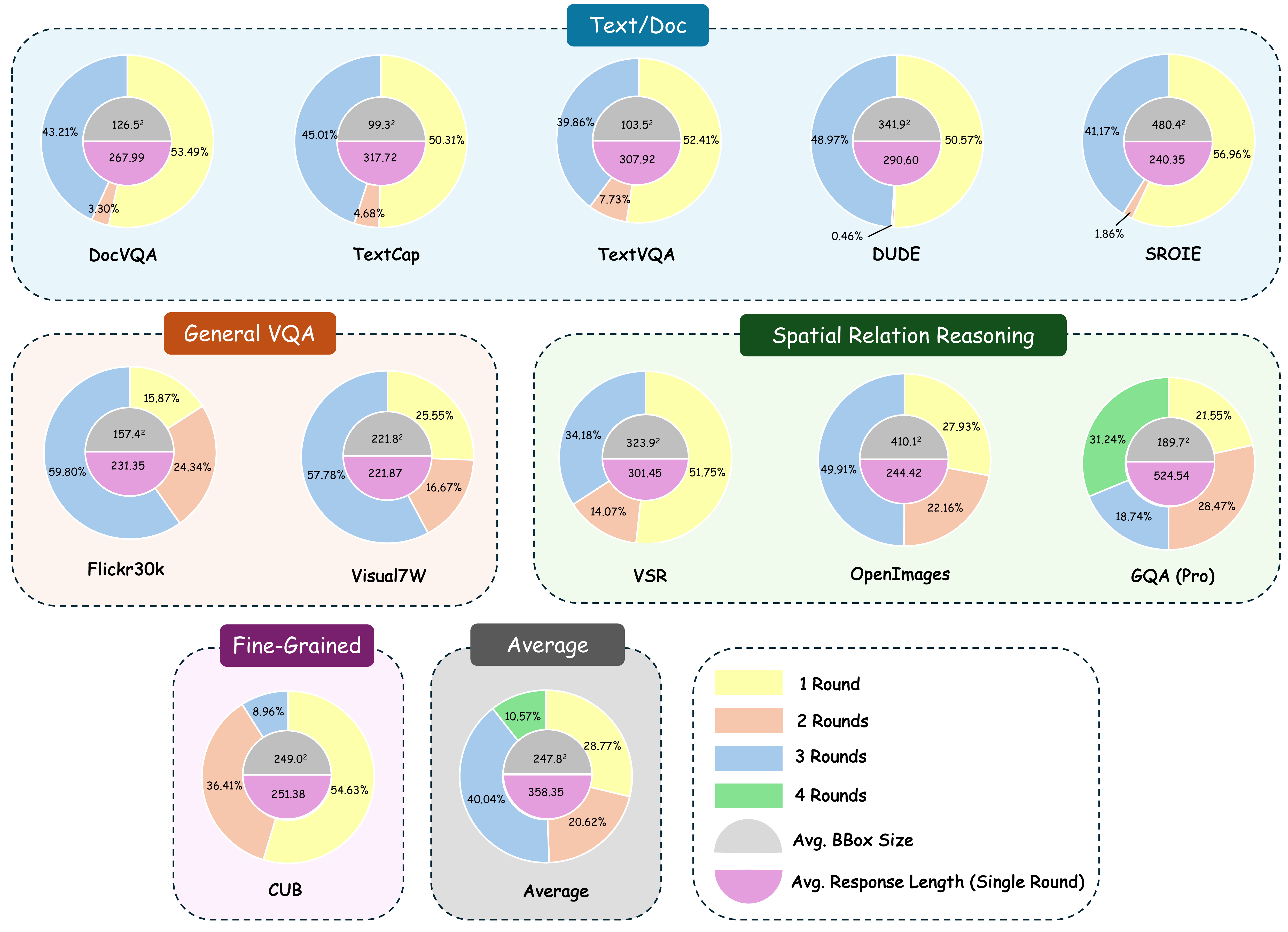}

  \caption{
  \textbf{Statistics of the proposed \datasetname{} dataset.}
  We report the distribution of CoT rounds (1–4), the average bounding-box size, and the average response length per round for each source dataset, showing that \datasetname{} offers rich multi-round supervision and consistently long, detailed reasoning steps across diverse domains.
}

  \label{fig:dataset_stat}
  
\end{figure*}

We visualize corpus statistics in \cref{fig:dataset_stat} and summarize coverage in \cref{tab:dataset}. RoIs are strongly biased toward small regions -- especially in text/doc tasks -- showing that answer-critical evidence often occupies only a minor part of the image ($\sim13.2\%$ on average), reinforcing the need for models to \emph{localize, zoom, and verify}. Most samples resolve in 2–3 rounds, with harder spatial or depth-augmented cases extending to 4, while easier ones naturally default to a single-step rationale. Notably, the \emph{per-round response length} remains consistently substantial, indicating detailed supervision rather than brief hints and surpassing prior process-level datasets~\cite{shao2024visualcot, zhang2025cof, qi2024cogcom}. Overall, \datasetname{} provides large-scale, explicit multi-round reasoning across diverse domains, complemented by a depth-aware subset (\datasetnamemax{}) that strengthens grounded 2D and ordinal-depth spatial reasoning.

\begin{table}[tb]
\caption{\textbf{Overview of the \datasetname{} dataset.} It spans four distinct domains and aggregates diverse source datasets, providing broad coverage of visual data styles while supplying spatial-aware, process-level supervision for robust visual chain-of-thought reasoning.}
\label{tab:dataset}
\centering
\resizebox{\textwidth}{!}{
\begin{tabular}{l|c|c|c|c|c}
\toprule
\textbf{Domain} & \textbf{Source Dataset} & \multicolumn{2}{c|}{\textbf{Train/Val Size}} & \textbf{GPT Model} & \textbf{Dataset Description}\\
\midrule
\multirow{5}{*}{\textbf{Text/Doc}} & TextVQA~\cite{singh2019towards} & 16k & 526 & 4.1-nano & Images with text \\
& TextCaps~\cite{sidorov2020textcaps} & 32k & 846 & 4.1-nano & Images with text \\
& DocVQA~\cite{mathew2021docvqa} & 50k & 846 & 4.1-nano & Doc Images \\
& DUDE~\cite{van2023document} & 11k & 559 &  4.1-nano & Doc Images \\
& SROIE~\cite{huang2019icdar2019} & 2k & 685 & 4.1-nano & Invoice Images \\
\midrule
\makecell[l]{\textbf{Fine-Grained}\\\textbf{Understanding}} & Birds-200-2011~\cite{Wah2011CUB_200_2011} & 10k & 491 & 4.1-nano & Images of birds \\
\midrule
\multirow{2}{*}{\textbf{General VQA}} & Flickr30k~\cite{plummer2015flickr30k} & 126k & 1455 & 4.1-nano & Images \\
 & Visual7W~\cite{zhu2016visual7w} & 30k & 994 & 4.1-nano & Images \\
\midrule
\multirow{3}{*}{\textbf{Spatial Relation Reasoning}} & VSR~\cite{Liu2022VisualSR} & 3k & 404 & 4.1-nano & Images \\
& GQA~\cite{hudson2019gqa} \textbf{(Pro)} & 165k & 978 & 4.1-mini & \makecell[c]{Images \textbf{(with spatial-aware} \\ \textbf{detailed reasoning steps)}} \\
& Open Images~\cite{kuznetsova2020open} & 43k & 944 & 4.1-nano & Images \\
\bottomrule
\end{tabular}%
}
\end{table}

\section{Enhancing MLLMs with CoT Reasoning Capabilities}

\begin{figure}[h!]

\centering
\includegraphics[width=0.95\linewidth]{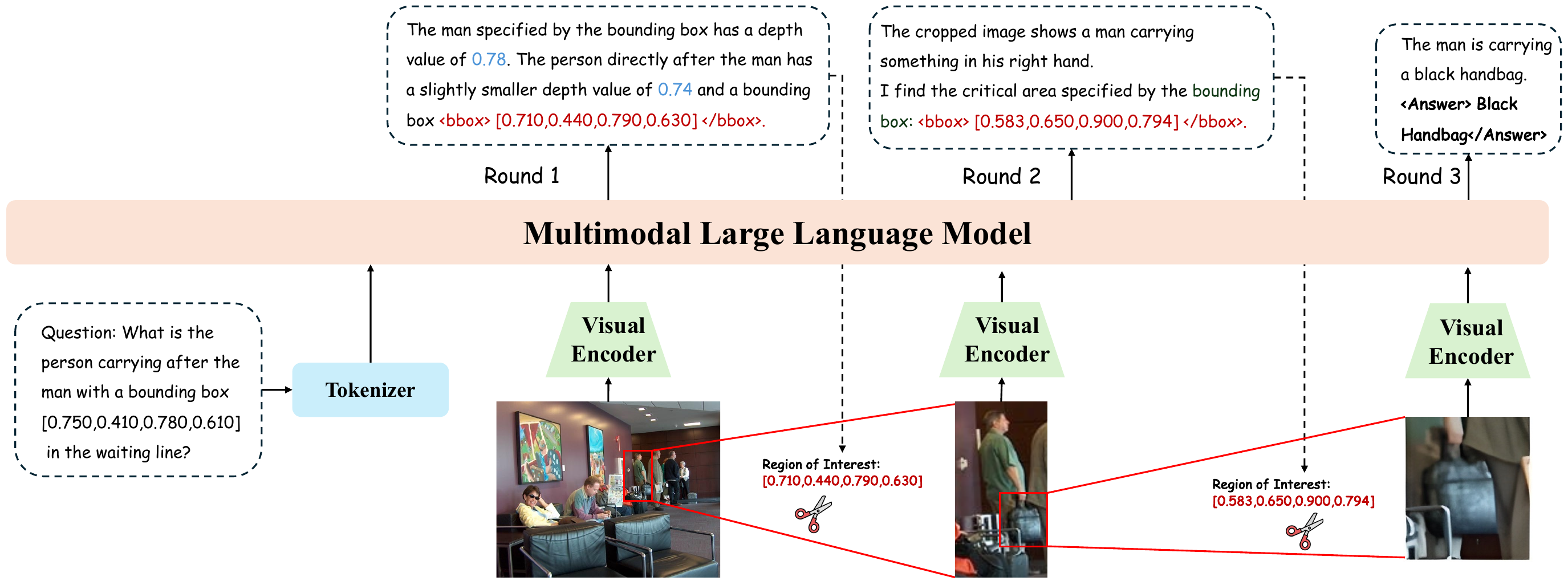}
\caption{\textbf{Overview of \datasetname{} paradigm}. The model iteratively processes the query by first generating a textual rationale and a bounding box for the next region of interest. It then crops the original image to this region, extracts new visual features, and appends them to the context to inform the next reasoning step, creating a zoom-and-verify sequence.
}
\label{fig:MLLM}
\end{figure}

\noindent\textbf{Formulation and Training.}
\label{sec:formulation-training}
Given an image $I$ and textual query $Q$, our model generates a multi-step reasoning process $Y=(a_0,a_1,\dots,a_T)$ to derive the final answer (\cref{fig:MLLM}). At step $t$, the action $a_t=(r_t,b_{t+1})$ consists of a textual reasoning snippet $r_t$ and a bounding box $b_{t+1}$ for the next region of interest. Generation is conditioned on prior actions and their visual inputs: the visual context at step $t$ is obtained by cropping $I$ with $b_t$ from the previous step, and we denote features by $\mathcal{V}(\text{crop}(I,b_t))$. The process is initialized with $b_0$ as the full image. The model auto-regressively outputs the tokens of each $a_t$ -- both the rationale and the serialized box coordinates -- based on the initial query and the full history of preceding visual and textual data:
\begin{equation} 
a_t \sim P_\theta(\cdot | Q, a_0, \dots, a_{t-1}, \mathcal{V}(\text{crop}(I, b_0)), \dots, \mathcal{V}(\text{crop}(I, b_t))). 
\end{equation}

We fine-tune this model on \datasetname{} via Supervised Fine-Tuning (SFT) using Qwen2.5-VL-7B~\cite{bai2025qwen25vl} as the base. During fine-tuning, we apply LoRA~\cite{hu2022lora} for efficient adaptation. The objective maximizes the likelihood of the ground-truth sequence $Y$ given $(I,Q)$ in a standard autoregressive manner, predicting the next token at each step. Concretely, we minimize the negative log-likelihood:
\begin{equation} 
\mathcal{L}(\theta) = - \sum_{(I, Q) \in \mathcal{D}} \sum_{t=1}^{|Y|} \log P_{\theta}(Y_t | I, Q, Y_{<t}), 
\label{eq:training_loss} 
\end{equation}
where $\mathcal{D}$ is the training set, $\theta$ the trainable parameters, and $Y_t$ the $t$-th token of $Y$. The sequence $Y$ is formed by serializing the multi-step CoT output, converting each $b_i$ into discrete tokens, so the model is trained end-to-end to produce both textual reasoning steps and serialized RoI coordinates for focusing visual attention.

\section{Experiments}
\label{sec:experiment}

\noindent\textbf{Training Details.}~~
We fine-tune Qwen2.5-VL-7B~\cite{bai2025qwen25vl} on \datasetname{} and \datasetnamemax{}. The baseline \datasetname{}-7B model is first trained for two epochs on \datasetname{} excluding \datasetnamemax{}, and \datasetnamemax{}-7B is then obtained by fine-tuning for one additional epoch on the full corpus including \datasetnamemax{}. We adopt a learning rate of $2\times10^{-5}$ for both the LLM backbone and the projector, and freeze the ViT encoder. To test backbone transferability, we additionally fine-tune InternVL-2.5-8B~\cite{zhu2025internvl3} on \datasetname{} for one epoch. Additional training configurations and ablations are provided in the Supplementary Materials.


\noindent\textbf{Benchmarks.}~~
We follow the Visual-CoT evaluation protocol~\cite{shao2024visualcot} and benchmark our models on its 11 source test sets (\cref{fig:dataset_stat}). As motivated in \cref{sec:intro}, we group tasks into four domains: text/document understanding, fine-grained recognition, general VQA, and spatial relational reasoning -- to capture both perceptual competence and multi-step spatial inference. To assess behavior beyond this evaluation suite, we additionally report results on MME~\cite{fu2023mme} and V*Bench~\cite{wu2024vstar}, two widely used LVLM benchmarks with short-answer or multiple-choice evaluation protocols. For automatic evaluation, we follow prior MLLM work~\cite{li2023videochat,luo2023valley,shao2024visualcot} and use GPT-4o-mini as an LLM-based judge to assign a scalar score in $[0,1]$ for each example. We run the judge five times and report averaged scores. The judge prompt, decoding setting, parser, scoring script, and human-calibration results are provided in the Supplementary Materials.


\begin{table}[tb]
\caption{\textbf{Comparison with state-of-the-art MLLMs} on Visual CoT benchmark. \textbf{Bold} indicates the best results and \underline{underline} indicates the second best results. The Average column is computed as the sample-size-weighted mean over all source test sets.}
\label{tab:comparison_internal}
\centering
\renewcommand{\arraystretch}{1.08}
\resizebox{\textwidth}{!}{%
\begin{tabular}{l|ccccc|c}
\toprule
\multirow{2}{*}{\textbf{MLLM}} & \multicolumn{5}{c|}{\textbf{Doc/Text}} & \textbf{Fine-grained} \\
\cmidrule(lr){2-6}\cmidrule(lr){7-7}
& DocVQA & TextCaps & TextVQA & DUDE & SROIE & Birds-200-2011 \\
\midrule
MiniGPTv2~\cite{chen2023minigpt} & 0.118 & 0.378 & 0.360 & 0.134 & 0.010 & 0.678 \\
VisCoT-7B~\cite{shao2024visualcot} & 0.476 & 0.675 & 0.775 & 0.386 & 0.470 & 0.559 \\
LLaVA-NeXT-8B~\cite{openllava1m} & 0.728 & 0.775 & 0.850 & 0.581 & 0.666 & 0.715 \\
InternVL-2.5-8B~\cite{zhu2025internvl3} & 0.846 & 0.829 & 0.907 & 0.716 & 0.907 & 0.747 \\
CoF-SFT-7B~\cite{zhang2025cof} & \underline{0.955} & \underline{0.867} & \underline{0.934} & \underline{0.813} & \underline{0.979} & 0.641 \\
Qwen2.5-VL-7B~\cite{bai2025qwen25vl} & \textbf{0.964} & \textbf{0.871} & \textbf{0.952} & \textbf{0.817} & \textbf{0.987} & 0.681 \\
\midrule
\textbf{\datasetname{}-7B} & 0.926 & 0.835 & 0.905 & 0.778 & 0.949 & \underline{0.792} \\
\textbf{\datasetnamemax{}-7B} & 0.928 & 0.847 & 0.922 & 0.786 & 0.966 & \textbf{0.831} \\
\midrule\midrule
\multirow{2}{*}{\textbf{MLLM}} & \multicolumn{2}{c|}{\textbf{General VQA}} & \multicolumn{3}{c|}{\textbf{Spatial Relation Reasoning}} & \multirow{2}{*}{\textbf{Average}} \\
\cmidrule(lr){2-3}\cmidrule(lr){4-6}
& Flickr30k & \multicolumn{1}{c|}{Visual7W} & GQA & Open Images & VSR & \\
\midrule
MiniGPTv2~\cite{chen2023minigpt} & 0.563 & \multicolumn{1}{c|}{0.635} & 0.656 & 0.615 & 0.626 & 0.452 \\
VisCoT-7B~\cite{shao2024visualcot} & 0.668 & \multicolumn{1}{c|}{0.558} & 0.631 & \textbf{0.822} & 0.614 & 0.614 \\
LLaVA-NeXT-8B~\cite{openllava1m} & 0.755 & \multicolumn{1}{c|}{0.703} & \textbf{0.736} & 0.559 & 0.647 & 0.705 \\
InternVL-2.5-8B~\cite{zhu2025internvl3} & 0.713 & \multicolumn{1}{c|}{0.681} & \underline{0.689} & 0.502 & \textbf{0.737} & 0.738 \\
CoF-SFT-7B~\cite{zhang2025cof} & 0.606 & \multicolumn{1}{c|}{0.686} & 0.674 & 0.503 & 0.657 & 0.737 \\
Qwen2.5-VL-7B~\cite{bai2025qwen25vl} & \underline{0.772} & \multicolumn{1}{c|}{0.690} & 0.651 & 0.498 & \underline{0.705} & 0.770 \\
\midrule
\textbf{\datasetname{}-7B} & 0.766 & \multicolumn{1}{c|}{\underline{0.695}} & 0.619 & 0.798 & 0.654 & \underline{0.787} \\
\textbf{\datasetnamemax{}-7B} & \textbf{0.777} & \multicolumn{1}{c|}{\textbf{0.698}} & 0.670 & \underline{0.805} & 0.654 & \textbf{0.802} \\
\bottomrule
\end{tabular}%
}
\end{table}

\begin{table}[t]
\centering
\caption{\textbf{Effect of inference format on the Visual-CoT evaluation suite.}
We evaluate the same \datasetnamemax{}-7B checkpoint with either the default multi-round visual-CoT format or a matched direct-QA prompt following the vanilla Qwen answer format. Domain scores and the overall average are computed as sample-size-weighted means. Direct-QA largely recovers Text/Doc performance, while multi-round CoT remains stronger on fine-grained and spatial reasoning. Full per-dataset results are provided in the Supplementary Materials.}
\label{tab:direct_qa_format_analysis}
\resizebox{1.0\linewidth}{!}{%
\begin{tabular}{lccccc}
\toprule
\textbf{Model / Prompt} & \textbf{Text/Doc} & \textbf{Fine-grained} & \textbf{General VQA} & \textbf{Spatial Rel.} & \textbf{Avg.} \\
\midrule
Qwen2.5-VL-7B, direct QA
& \textbf{0.920} & 0.681 & 0.739 & 0.598 & 0.770 \\
\datasetnamemax{}-7B, multi-round CoT
& 0.892 & \textbf{0.831} & \textbf{0.745} & \textbf{0.722} & \textbf{0.802} \\
\datasetnamemax{}-7B, direct QA
& 0.916 & 0.709 & 0.735 & 0.599 & 0.769 \\
\bottomrule
\end{tabular}%
}
\end{table}

\begin{table}[t]
\caption{\textbf{Detection performance} on the \datasetnamemax{} held-out suite. The ground-truth bounding boxes used for computing the metric are the final-round CoT region-of-interest bounding boxes annotated in \datasetnamemax{}.}
\centering
 \resizebox{0.9\linewidth}{!}{%
\begin{tabular}{c|c|c|c|c|c}
\toprule
  \textbf{Metric} & MiniGPTv2  & LLaVA-NeXT & InternVL-2.5  & \textbf{\datasetname{}} & \textbf{\datasetnamemax{}}  \\
  \midrule
   \textbf{IoU@0.5 $\uparrow$}& 0.14 & 0.29 & 0.08 & 0.27 & \textbf{0.34}  \\
   \textbf{IoU@0.75 $\uparrow$} & 0.06 & 0.19 & 0.03 & 0.13 & \textbf{0.23} \\
  \bottomrule
\end{tabular}
}

\label{tab:detection}
\end{table}

\begin{table}[t]
\caption{\textbf{Ordinal-depth and RoI grounding performance} on the \datasetnamemax{} held-out suite. Depth error is computed from monocular-depth-derived ordinal cues and does not indicate metric 3D reconstruction accuracy.}
\label{tab:grounding}
\centering
 \resizebox{0.85\linewidth}{!}{%
\begin{tabular}{c|c|c|c|c}
\toprule
   & LLaVA-NeXT & InternVL-2.5  & Qwen2.5-VL & \textbf{\datasetnamemax{}}  \\
  \midrule
   \textbf{Grounded Ratio $\uparrow$} & 0.039 & 0.011 & 0.035 & \textbf{0.276}  \\
   \textbf{BBox (IoU) $\uparrow$} & 0.207 & 0.214 & 0.115 & \textbf{0.278} \\
   \textbf{Depth (Abs Diff) $\downarrow$} & 0.394 & 0.290 & 0.294 & \textbf{0.266} \\
  \bottomrule
\end{tabular}
}

\end{table}

\subsection{Comparison with State-of-the-art MLLMs}
\label{sec: comparison}

\noindent\textbf{Comparison on Visual-CoT-Benchmark.}~~
As shown in \cref{tab:comparison_internal}, our models achieve the strongest overall performance among the compared open-source MLLMs. \datasetnamemax{}--7B obtains the highest average score (0.802), improving over its Qwen2.5-VL-7B backbone (0.770) and outperforming InternVL-2.5-8B, LLaVA-NeXT-8B, and VisCoT-7B. The gains are most pronounced on fine-grained recognition (Birds-200-2011: $0.681 \rightarrow 0.831$) and spatial relation reasoning (OpenImages: $0.498 \rightarrow 0.805$), which are the settings most aligned with our RoI-grounded, global-to-local supervision.

\noindent\textbf{Error Analysis on Doc/Text Tasks.}
We observe a drop on Doc/Text benchmarks under the default multi-round CoT inference compared to the Qwen2.5-VL-7B base model. As shown in \cref{tab:direct_qa_format_analysis}, this drop is largely format-dependent: when the same \datasetnamemax{} checkpoint is evaluated with a matched direct-QA prompt, Text/Doc performance nearly recovers to the Qwen baseline. This suggests that the degradation is not an intrinsic loss of text/document understanding ability, but mainly arises from evaluating verbose RoI-grounded reasoning traces under short-answer QA protocols. At the same time, multi-round CoT remains preferable for fine-grained and spatial reasoning, where localized visual inspection is central to the task.

\noindent\textbf{Evaluation on the \datasetnamemax{} Benchmark.}~~
The \datasetnamemax{} benchmark evaluates two key abilities: (i) accurate RoI localization and (ii) explicit 3D grounding (grounded ratio, average IoU, and depth error). To ensure fair comparison and mitigate evaluation bias, all baselines are explicitly prompted to output bounding boxes using their native grounding templates. 
As shown in~\cref{tab:detection} and \cref{tab:grounding}, our models achieve the best RoI localization and stronger alignment with ordinal-depth cues. These results support the effectiveness of depth-informed process supervision, while we emphasize that the depth signal is pseudo-depth supervision rather than metric 3D ground truth.

\noindent\textbf{Generalization and the Evaluation Format Mismatch.}~~ 
External benchmarks such as MME and V* rely on short-answer or multiple-choice parsing, whereas our default inference uses verbose multi-round CoT with explicit RoI coordinates. To separate output-format mismatch from capability regression, we evaluate the same checkpoints under a matched direct-QA protocol without retraining. As shown in \cref{tab:comparison external}, direct-QA inference largely recovers MME and V* performance, while multi-round CoT remains stronger on the Visual-CoT benchmark. This indicates that the external drop mainly comes from protocol mismatch, and that \datasetname{} should be viewed as a specialization for RoI-grounded multi-step visual reasoning rather than a replacement for direct-QA inference in all settings.

\begin{table}[t]
\caption{Second-backbone transfer.}
\label{tab:internvl_transfer}
\centering
\resizebox{0.98\linewidth}{!}{%
\begin{tabular}{l|ccccc}
\toprule
\textbf{Model} & \textbf{Text/Doc} & \textbf{Fine-grained} & \textbf{General VQA} & \textbf{Spatial Rel.} & \textbf{Overall} \\
\midrule
InternVL-2.5-8B
& \textbf{0.842} & 0.747 & \textbf{0.700} & 0.621 & 0.738 \\
InternVL-2.5-8B + \datasetname{}
& 0.815 & \textbf{0.823} & 0.684 & \textbf{0.669} & \textbf{0.740} \\
\bottomrule
\end{tabular}}
\end{table}

\noindent\textbf{Second-backbone Transfer.}~~
As shown in \cref{tab:internvl_transfer}, fine-tuning InternVL-2.5-8B on \datasetname{} for one epoch produces a similar specialization pattern: fine-grained and spatial reasoning improve, while Doc/Text performance is less favorable under the multi-round format. This suggests that the effect of \datasetname{} is not specific to the Qwen2.5-VL backbone.

\subsection{Ablation Study}
\label{sec:ablation_study}
\cref{tab: ablation} isolates the contributions of our two dataset variants and the ``zoom-in when needed'' strategy. Training on \datasetname{} improves over the Qwen baseline, particularly on relation reasoning and fine-grained tasks, indicating that multi-round spatial supervision provides substantial benefit. Incorporating \datasetnamemax{} further strengthens general VQA, relation reasoning, and fine-grained recognition, reflecting the value of higher-fidelity rationales and depth-aware cues. Adding the adaptive zoom-in mechanism yields the best overall performance, boosting fine-grained and relation-heavy benchmarks while largely preserving Doc/Text accuracy. These results confirm that both richer supervision and selective zooming contribute meaningfully to stronger and more balanced visual reasoning.

\begin{table}[t]
\centering
\begin{minipage}[t]{0.48\linewidth}
    \caption{\textbf{Effect of inference protocol on external benchmarks.}
    DQA = direct QA; MR = multi-round CoT; VR-Pro = \datasetnamemax{}.}
    \label{tab:comparison external}
    \centering
    \resizebox{\linewidth}{!}{%
    \begin{tabular}{@{}llccc@{}}
    \toprule
    \textbf{Model} & \textbf{Prot.} & \textbf{V-CoT} & \textbf{MME} & \textbf{V*} \\
    \midrule
    Qwen2.5 & DQA
    & 0.770 & \textbf{0.861} & \textbf{0.791} \\
    VR-Pro & MR
    & \textbf{0.802} & 0.777 & 0.603 \\
    VR-Pro & DQA
    & 0.769 & 0.856 & \textbf{0.791} \\
    \midrule
    InternVL & DQA
    & 0.738 & 0.848 & 0.597 \\
    InternVL+VR & MR
    & 0.740 & 0.765 & 0.539 \\
    InternVL+VR & DQA
    & -- & 0.843 & 0.598 \\
    \bottomrule
    \end{tabular}%
    }
\end{minipage}
\hfill
\begin{minipage}[t]{0.48\linewidth}
    \caption{\textbf{Human evaluation (1--5).} AA = Answer Accuracy; GF = Grounded Faithfulness; SCS = Stepwise Clarity \& Sufficiency.}
    \label{tab:userstudy}
    \centering
    \resizebox{\linewidth}{!}{%
    \begin{tabular}{@{}lcccc@{}}
    \toprule
    \textbf{Method} & \textbf{AA} & \textbf{GF} & \textbf{SCS} & \textbf{Mean} \\
    \midrule
    MiniGPTv2~\cite{chen2023minigpt}  & 2.37 & 1.94 & 1.88 & 2.06 \\
    VisCoT-7B~\cite{shao2024visualcot} & 2.83 & 2.58 & 2.34 & 2.58 \\
    LLaVA-NeXT-8B~\cite{openllava1m} & 3.52 & 2.93 & 3.08 & 3.18 \\
    InternVL-2.5-8B~\cite{zhu2025internvl3} & 3.87 & 3.32 & 3.24 & 3.48 \\
    \midrule
    \textbf{\datasetname{}-7B} & 4.07 & 4.18 & 4.12 & 4.12 \\
    \textbf{\datasetnamemax{}-7B} & \textbf{4.19} & \textbf{4.46} & \textbf{4.37} & \textbf{4.34} \\
    \bottomrule
    \end{tabular}%
    }
\end{minipage}
\end{table}

\begin{table}[t]
\caption{\textbf{Ablation study} on dataset selection and zoom-in strategy. \textbf{Baseline} refers to \datasetname{}, \textbf{Pro} refers to \datasetnamemax{}, and \textbf{AZ} refers to the \textbf{A}daptive \textbf{Z}oom-In strategy. Domain scores and the overall average are computed as sample-size-weighted means.}
\label{tab: ablation}
\centering
\resizebox{0.95\linewidth}{!}{%
\begin{tabular}{ccc|c|c|c|c|c}
\toprule
\textbf{Baseline} & \textbf{Pro} & \textbf{AZ} & Doc/Text & General VQA & Relation Reasoning & Fine-grained & Average \\
\midrule
& & & \textbf{0.920} & 0.739 & 0.598 & 0.681 & 0.770 \\
\ding{52} & & & 0.864 & 0.744 & 0.678 & 0.798 & 0.777 \\
\ding{52} & & \ding{52} & 0.882 & 0.738 & 0.705 & 0.792 & \underline{0.789} \\
\ding{52} & \ding{52} & & 0.856 & \textbf{0.750} & 0.693 & \underline{0.809} & 0.780 \\
\ding{52} & \ding{52} & \ding{52} & \underline{0.892} & \underline{0.745} & \textbf{0.722} & \textbf{0.831} & \textbf{0.802} \\
\bottomrule
\end{tabular}%
}
\end{table}

\subsection{User Study}
\label{sec:user_study}
We conducted a blinded study with 30 raters on 20 sampled items per method, evaluating Answer Accuracy (AA), Grounded Faithfulness (GF), and Stepwise Clarity \& Sufficiency (SCS). As shown in \cref{tab:userstudy}, models trained on \datasetname{} achieve clear gains in GF and SCS -- reflecting tighter RoIs and more coherent global-to-local chains -- which also boosts AA. \datasetnamemax{} further strengthens all three criteria, with raters noting more reliable grounding and more complete reasoning steps. These results confirm that multi-round, depth-aware supervision substantially improves the quality and faithfulness of visual reasoning.

\subsection{Discussion and Limitations}
\label{sec: discussion}

\noindent\textbf{Data Quality, Ordinal-Depth Noise, \& Human Validation.}~~
We conduct a stratified blind human audit on 2,200 examples, sampling 200 examples from each source dataset and aggregating the results into four domains. The audit shows high answer and RoI quality overall: 99.1\% answer consistency, 98.5\% target containment, and 95.0\% RoI tightness, with 86.5\% rationale necessity and 86.5\% rationale faithfulness. The main weakness appears in Text/Doc rationales, where local crops can miss global layout cues required for document reasoning. Regarding depth supervision, monocular depth provides useful ordinal cues but remains noisy and should not be interpreted as metric 3D ground truth. Full per-domain audit results with Wilson 95\% confidence intervals are provided in the Supplementary Materials. We also validate the GPT-4o-mini judge against human ratings and provide the prompt, parser, and scoring script for reproducibility.

\noindent\textbf{Inference Cost \& SFT vs. RL.}~~
Iterative zoom-and-verify increases inference latency, but this adaptive trade-off allocates more computation to visually complex queries. While Reinforcement Learning (RL) could, in principle, learn such cropping policies, doing so from sparse final-answer rewards is difficult and may encourage reward hacking. Large-scale SFT datasets like \datasetname{} therefore provide useful RoI-grounded trajectory supervision for future multimodal RL or agentic training.

\begin{figure}[h!]
\centering
\includegraphics[width=0.95\linewidth]{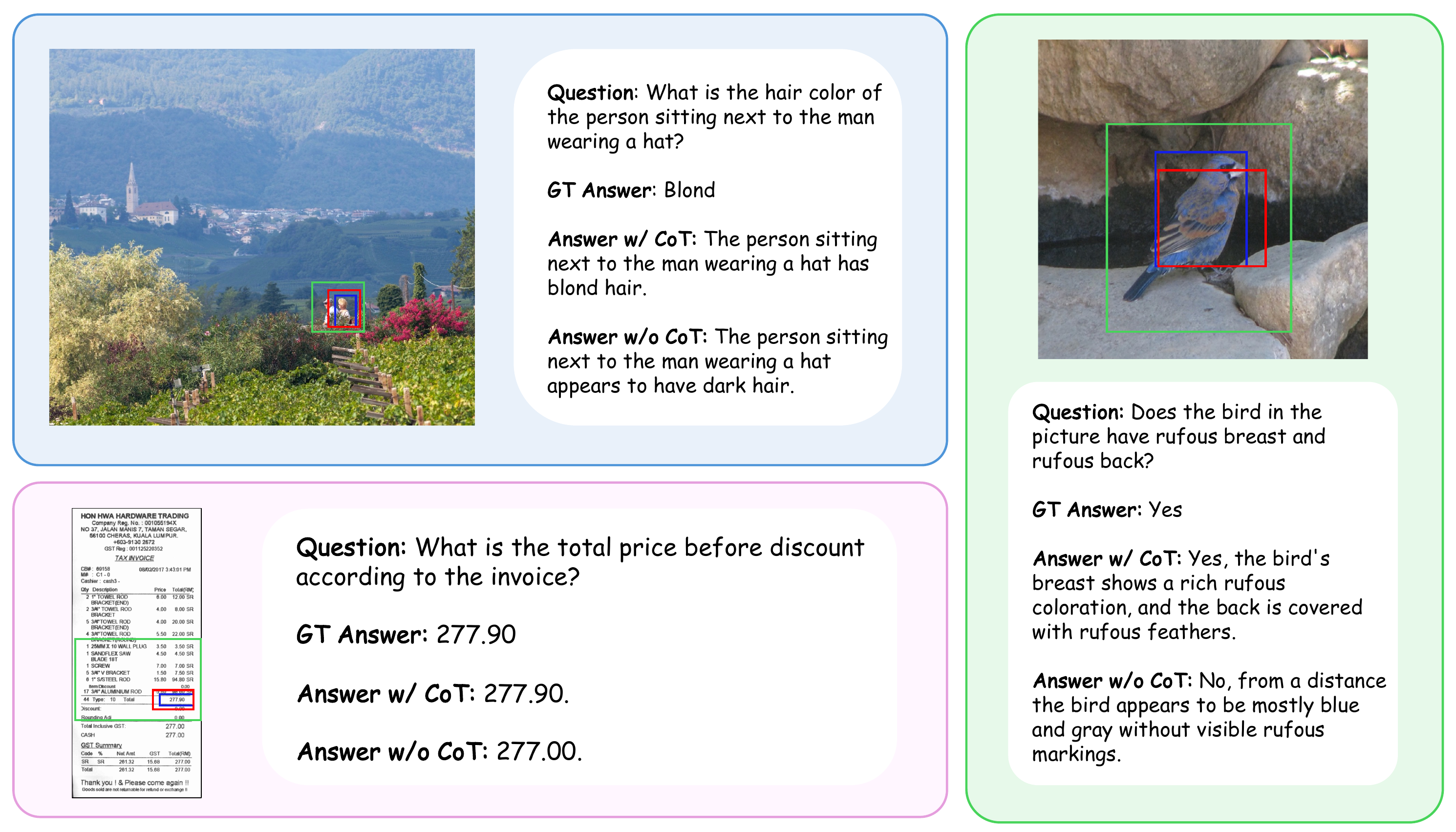}
\caption{Visualization results of \datasetname{} to illustrate the difference between various inference modes. Model-generated bounding boxes are shown in \textcolor{green}{green} (first-round) and \textcolor{red}{red} (second-round), while ground truth (GT) bounding boxes are in \textcolor{blue}{blue}. Best viewed in color and zoomed in.
}
\label{fig:demo}
\end{figure}

\subsection{Visualization}
\label{sec:visualization}
\cref{fig:demo} illustrates our model under CoT and non-CoT inference modes. The CoT mode progressively localizes critical regions and integrates information from both the original and zoomed-in views, leading to more reliable grounded reasoning and answer quality.


\noindent\textbf{Cascading Localization Failures.}~~
Our dynamic \textit{zoom-and-verify} paradigm introduces a mechanistic vulnerability: \textit{cascading localization errors}. Since subsequent steps are conditioned on previous cropped features, an initially misaligned RoI can trap the model's perception in an irrelevant region. Unlike single-pass MLLMs that retain global context, our iterative mechanism lacks a ``zoom-out'' or backtracking policy to recover from early failures. Developing self-correcting rollback mechanisms remains an important direction for future work.

\section{Conclusion}
\label{sec:conclusion}
In summary, we address key limitations of existing visual CoT -- insufficient scale, lack of multi-round supervision, and weak depth-aware spatial reasoning -- by introducing \textbf{\datasetname{}} and \textbf{\datasetnamemax{}} with compact global-to-local rationales, RoI boxes, and, in \datasetnamemax{}, pseudo-depth and segmentation cues derived from monocular depth estimation and semantic segmentation. Spanning text/doc, fine-grained, general VQA, and spatial relational tasks, these resources provide rich process-level supervision for faithful and spatially grounded inference. Building upon them, we establish a held-out evaluation suite for assessing stepwise reasoning, RoI localization, and ordinal-depth-augmented spatial reasoning. Experiments across different MLLM backbones show consistent gains in fine-grained recognition, spatial reasoning, RoI grounding, and human-rated faithfulness. We hope this dataset family and evaluation suite offer a useful foundation for advancing spatially aware multimodal reasoning.

\section*{Acknowledgments}
This work was supported in part by the Institute for Advanced Computational Science at Stony Brook University for providing the computational resources and support that contributed to the research presented in this publication. The authors thank Mr. Yuan Qing at Boston University for his careful reading of the manuscript and constructive suggestions.

\bibliographystyle{splncs04}
\bibliography{main}
\appendix
\clearpage
\section*{ \huge Supplementary Material\vspace{1em}}
\label{appendix: supplementary material}

\section*{Overview}
This appendix is organized as follows:

\begin{itemize}
\item \cref{appendix:implementation details}:
Training configuration and implementation details for \datasetname{}--7B and \datasetnamemax{}--7B.

\item \cref{appendix:data generation}:  
Full data-generation pipelines.
\begin{itemize}
    \item \cref{appendix:visreason}: Construction of the 2D spatial CoT dataset \datasetname{}.
    \item \cref{appendix:visreason-pro}: Construction of the depth-aware dataset \datasetnamemax{}.
\end{itemize}

\item \cref{appendix:evaluation}:  
Evaluation protocol on Visual-CoT-Benchmark.

\item \cref{appendix:additional results}:  
Additional evaluation results.
\begin{itemize}
    \item \cref{appendix:direct_qa_format_analysis}: Full Direct-QA format analysis.
    \item \cref{appendix:second_backbone_validation}: Second-backbone validation on InternVL-2.5-8B.
    \item \cref{appendix:source_removal_ablation}: Source-held-out / source-removal ablation.
    \item \cref{appendix:human_audit}: Human audit of data quality.
    \item \cref{appendix:gpt_judge_human_calibration}: Human calibration of the GPT-based evaluator.
\end{itemize}

\item \cref{appendix:visreason-pro-eval}:  
Evaluation protocol on \datasetnamemax{}.
\begin{itemize}
    \item \cref{appendix:roi-eval}: RoI localization evaluation.
    \item \cref{appendix:3d-grounding-eval}: 3D grounding evaluation.
\end{itemize}

\item \cref{appendix:prompts}:  
All prompts used in dataset generation, refinement, grounding, and grading.

\item \cref{appendix:more data}:  
Additional dataset examples.

\item \cref{appendix:more inference exmaples}:  
Additional inference examples.
\end{itemize}

\section{Implementation Details}

\label{appendix:implementation details}

We train both \datasetname{}-7B and \datasetnamemax{}-7B models on a cluster equipped with four NVIDIA H200 GPUs (144 GB memory each). Training is implemented in PyTorch using the HuggingFace LLaMA-Factory framework, with distributed execution based on DeepSpeed ZeRO-2 with CPU offloading. FlashAttention-2 is enabled for efficient attention computation, and KV-cache is disabled to increase training stability. All experiments use bf16 precision.

We adopt LoRA-based parameter-efficient finetuning with a rank of 32 and apply LoRA to all trainable modules. The vision tower of Qwen2.5-VL-7B is frozen throughout training, while the multi-modal projector and language model remain fully trainable. Gradient checkpointing (non-reentrant mode) is activated to reduce memory overhead, allowing a global batch size of 96 on 4 GPUs (per-device batch size 3 × gradient accumulation 4 × 4 GPUs).

Images are dynamically resized with a minimum of $112 \times 112$ pixels (12,544 pixels) and a maximum of $512 \times 512$ pixels (262,144 pixels). The maximum sequence length is set to 8{,}192 tokens, and sample packing is enabled to improve training efficiency. Dataset preprocessing is parallelized using 128 workers, and data loading uses 12 workers per GPU with pinned memory, batch prefetching (factor 6), and persistent worker reuse.

Optimization is performed using the fused AdamW optimizer with a cosine learning rate schedule, a peak learning rate of $2.0\times 10^{-5}$, warmup ratio of 1\%, gradient clipping at 1.0, and dropout disabled. All unused parameters are removed during forward passes to reduce memory waste, and distributed training uses \texttt{ddp\_find\_unused\_parameters = false}. Checkpoints are saved every 50 steps in \texttt{.safetensors} format with a maximum history of three versions.

All evaluations and ablations are performed on a single NVIDIA A800 GPU (80 GB memory) under bf16 precision.

\cref{tab:training-config-pro} and \cref{tab:training-config-pro-mx} summarize the main hyperparameters used for \datasetname{}-7B and \datasetnamemax{}-7B, respectively.

\begin{table}[t]
\centering
\caption{Detailed training configuration for \datasetname{}.}
\label{tab:training-config-pro}
\footnotesize 
\begin{tabular}{lc}
\toprule
\textbf{Configuration} & \datasetname{}-7B \\
\midrule
\textbf{Hardware} & 4$\times$ H200 (144 GB) \\
Precision & bf16 \\
Distributed engine & DeepSpeed ZeRO-2 + CPU offload \\
Flash attention & FlashAttention-2 \\
LoRA rank & 32 (all trainable modules) \\
Frozen modules & Vision tower only \\
\midrule
\textbf{Dataset / Input} & \\
Dataset Use & \datasetname{} \\
Cutoff length (tokens) & 8192 \\
Image max pixels & 262,144 (512$\times$512) \\
Image min pixels & 12,544 (112$\times$112) \\
Packing & Enabled \\
\midrule
\textbf{Training} & \\
Base model & Qwen2.5-VL-7B-Instruct \\
Global batch size & 96 \\
Per-device batch size & 3 \\
Gradient accumulation & 4 \\
Epochs & 2 \\
Optimizer & AdamW (fused) \\
Learning rate & 2.0$\times 10^{-5}$ \\
Scheduler & Cosine, warmup 1\% \\
Gradient clipping & 1.0 \\
Gradient checkpointing & Enabled (non-reentrant) \\
Use cache (KV) & Disabled \\
Group-by-length & Enabled \\
Drop last batch & True \\
\midrule
\textbf{Dataloader} & \\
Preprocessing workers & 128 \\
Dataloader workers & 12 $\times$ 4 = 48 \\
Pin memory & True \\
Prefetch factor & 6 \\
Persistent workers & True \\
\midrule
\textbf{Saving} & \\
Save every steps & 50 \\
Max checkpoints & 3 \\
Format & safetensors \\
\bottomrule
\end{tabular}
\end{table}

\begin{table}[t]
\centering
\caption{Detailed training configuration for \datasetnamemax{}.}
\label{tab:training-config-pro-mx}
\footnotesize 
\begin{tabular}{lc}
\toprule
\textbf{Configuration} & \datasetnamemax{}-7B \\
\midrule
\textbf{Hardware} & 4$\times$ H200 (144 GB) \\
Precision & bf16 \\
Distributed engine & DeepSpeed ZeRO-2 + CPU offload \\
Flash attention & FlashAttention-2 \\
LoRA rank & 32 (all trainable modules) \\
Frozen modules & Vision tower only \\
\midrule
\textbf{Dataset / Input} & \\
Dataset Use & \datasetnamemax{} \\
Cutoff length (tokens) & 8192 \\
Image max pixels & 262,144 (512$\times$512) \\
Image min pixels & 12,544 (112$\times$112) \\
Packing & Enabled \\
\midrule
\textbf{Training} & \\
Base model & \datasetname{}-7B \\
Global batch size & 96 \\
Per-device batch size & 3 \\
Gradient accumulation & 4 \\
Epochs & \textbf{1} \\
Optimizer & AdamW (fused) \\
Learning rate & 2.0$\times 10^{-5}$ \\
Scheduler & Cosine, warmup 1\% \\
Gradient clipping & 1.0 \\
Gradient checkpointing & Enabled (non-reentrant) \\
Use cache (KV) & Disabled \\
Group-by-length & Enabled \\
Drop last batch & True \\
\midrule
\textbf{Dataloader} & \\
Preprocessing workers & 128 \\
Dataloader workers & 12 $\times$ 4 = 48 \\
Pin memory & True \\
Prefetch factor & 6 \\
Persistent workers & True \\
\midrule
\textbf{Saving} & \\
Save every steps & 50 \\
Max checkpoints & 3 \\
Format & safetensors \\
\bottomrule
\end{tabular}
\end{table}

\section{Data Generation Details}
\label{appendix:data generation}

\subsection{Construction of the \datasetname{} Dataset}
\label{appendix:visreason}

\datasetname{} is derived from the \emph{regular} portion of the Visual-CoT dataset~\cite{shao2024visualcot}. The original split provides, for each sample, an image, a natural-language question, a short answer, a long-form explanation, and a set of object annotations with semantic labels and \texttt{xywh}-formatted bounding boxes. However, the regular split does not contain any process-level reasoning, intermediate region-of-interest (RoI) annotations, or multi-step spatial justification. To obtain explicit spatial chain-of-thought (CoT) supervision, we enrich every sample with multi-round descriptions, rationales, and refined RoIs.

\paragraph{Round-1 expansion.}
For each image--QA pair, we first perform a single-round expansion that prompts a vision--language model (VLM) with the full-resolution image, the question, and the short answer. The model returns three components: (i) a compact description of the scene, (ii) a normalized RoI $\hat{A}_1 \in [0,1]^4$, and (iii) a brief rationale that justifies the predicted region. We convert $\hat{A}_1$ to pixel coordinates and apply an adjustment operation that guarantees coverage of the ground-truth bounding box $B^\star$ while respecting image boundaries.

\paragraph{Multi-round iterative refinement.}
To capture finer spatial relations, we apply an adaptive multi-round zoom-in strategy. After Round~1, if the predicted RoI remains significantly larger than the GT box (\textit{i.e.}, $\mathrm{Region}(A_t) > N \cdot \mathrm{Region}(B^\star)$), we crop the current RoI from the image and prompt the VLM again using only the cropped region. The model produces a new triplet $(\mathrm{desc}_{t+1}, \hat{A}_{t+1}, \mathrm{reason}_{t+1})$, where $\hat{A}_{t+1}$ is predicted in local coordinates. This box is mapped back to the global system, adjusted to cover $B^\star$, and used as the basis for the next iteration. The refinement continues for at most $R_{\max}=3$ rounds or until the region ratio constraint is satisfied.

\paragraph{Final justification.}
After the refinement terminates, the last cropped view is used to obtain a single-sentence final justification explicitly referencing the GT object. All intermediate results are stored as a multi-round chain:
\[
  \{(\mathrm{desc}_t, A_t, \mathrm{reason}_t)\}_{t=1}^{T}.
\]
This chain reflects a progressively localized reasoning process aligned with the final answer.

\paragraph{Optional consistency enhancement.}
To improve annotation quality, we perform an additional refinement pass that validates bounding boxes across rounds, corrects out-of-bound predictions, removes implausible RoIs, and ensures consistent local-to-global coordinate transformations. This results in more stable multi-round traces, especially in complex or cluttered scenes.

\paragraph{Final output.}
Each \datasetname{} sample contains scene descriptions, rationales, and refined RoIs at every round, together with a final justification. The result is a large-scale 2D spatial CoT dataset derived from Visual-CoT, providing explicit stepwise grounding signals for training multimodal reasoning models.

\subsection{Construction of the \datasetnamemax{} Dataset}
\label{appendix:visreason-pro}

\datasetnamemax{} builds upon the GQA-derived portion of Visual-CoT, which contains richer annotations than the regular split, including object categories, object-level bounding boxes, and fine-grained relational metadata. Nevertheless, these annotations remain purely 2D: neither GQA nor Visual-CoT provide depth maps, ordinal depth relationships, or 3D-aware QA pairs. \datasetnamemax{} augments this data with pseudo-3D cues and depth-sensitive spatial CoT traces.

\paragraph{Pseudo-3D annotation.}
To obtain depth- and geometry-aware signals for \datasetnamemax{}, each GQA image is augmented with both monocular depth and semantic segmentation derived from strong foundation models. Depth is estimated using \emph{Depth-Anything~V2}~\cite{yang2024depthanythingv2}, a state-of-the-art transformer-based monocular depth estimator that produces a dense depth field $D \in [0,1]^{H \times W}$ after min--max normalization. In parallel, we compute instance-level semantic segmentation using the \emph{Mask2Former} universal segmentation model~\cite{cheng2022masked}, based on a Swin-L large backbone, which yields per-pixel semantic labels and multi-object instance masks.

To further improve spatial consistency, especially for small or fragmented regions, we perform depth-guided region merging: small connected components in the semantic map are merged into adjacent regions whose mean depth differs by less than a threshold (0.15), ensuring geometrically coherent segmentation. In addition, object-level masks are refined using SAM2~\cite{ravi2024sam2}, from which high-quality binary masks are obtained for all sufficiently large objects.

For each detected object, we compute: (i) a segmentation-aligned bounding box, (ii) the robust depth estimate (median depth over the mask, corrected by depth statistics at anchor points), and (iii) an ordinal depth ranking among all objects in the image. This yields a structured object list
\[
    \mathcal{O} = \{(c_i, [x_1,y_1,x_2,y_2]_i, d_i)\}_{i=1}^{K},
\]
where $c_i$ denotes the semantic category, and $d_i$ is the object's ordinal depth in the global scene order. These pseudo-3D cues enable our generator to compose questions involving coupled 2D and depth relations (\emph{left of}, \emph{above}, \emph{in front of}, \emph{behind}), and support multi-round 3D-grounded reasoning in \datasetnamemax{}.

\paragraph{3D-aware QA generation.}
To introduce depth reasoning, we instruct a VLM to generate new QA pairs that jointly involve 2D spatial relations (\emph{left of}, \emph{above}) and depth relations (\emph{in front of}, \emph{behind}). The model receives the image and the enriched object list~$\mathcal{O}$ and produces a depth-aware question, a short answer, a long answer, and a GT bounding box for the referenced target object. The output is converted into pixel coordinates and normalized to match the style of the Visual-CoT annotations. This stage produces 3D-aware samples tightly aligned with the geometric structure of the scene.

\paragraph{Multi-round 3D-aware CoT refinement.}
We construct depth-sensitive multi-round CoT traces using a refinement process similar to the \datasetname{} pipeline but extended to incorporate 3D cues. At each round, the prompt includes the cropped image region and the subset of objects whose boxes fall inside the crop. Their depth values are re-normalized locally, ensuring that ordinal ordering remains consistent under cropping. The VLM predicts a description, a local RoI, and a rationale that may reference both spatial and depth relations. The predicted box is mapped back to global coordinates, adjusted to ensure coverage of the GT box, and passed to the next round. This refinement proceeds for up to $R_{\max}=4$ rounds or until the region ratio constraint is satisfied.

\paragraph{Grounding-aware rationale augmentation.}
To maintain alignment between textual reasoning and pixel evidence, we optionally augment the model-generated rationales by explicitly mentioning object identities, bounding boxes, and ordinal depths when they appear in the reasoning. This produces richer supervision signals that couple linguistic explanations with spatial and depth structure.

\paragraph{Final output.}
Each \datasetnamemax{} entry contains: (i) a depth-sensitive question, (ii) the short and long answers, (iii) an object list with segmentation-derived bounding boxes and ordinal depths, and (iv) a multi-round chain of $(\mathrm{desc}_t, A_t, \mathrm{reason}_t)$ triplets. A distilled single-round variant is also provided, compressing the entire CoT trajectory into one step. The resulting dataset forms a high-fidelity 2D–3D spatial reasoning corpus suitable for evaluating depth-aware multimodal models.

\clearpage

\begin{algorithm}[t]
\caption{\datasetname{}: Multi-round 2D Spatial CoT Generation}
\label{alg:visreason}
\begin{algorithmic}[1]
\Require Image $I$, question $q$, short answer $a$, GT box $B^\star$; max rounds $R_{\max}{=}3$; area ratio threshold $N{=}2$; large-object threshold $\tau_{\text{large}}$; generator $\mathcal{G}$
\Ensure Multi-round chain $\{(\mathrm{desc}_t, \mathrm{RoI}_t, \mathrm{reason}_t)\}_{t=1}^{T}$ and a final justification

\State $(W,H) \gets \text{size}(I)$
\State Compute image area $A_{\text{img}} \gets W \cdot H$ and GT area $A_{\text{GT}} \gets \mathrm{Region}(B^\star)$
\If{$A_{\text{GT}} / A_{\text{img}} \ge \tau_{\text{large}}$}
    \State Prompt $\mathcal{G}$ with $(I, q, a)$ to obtain $\mathrm{desc}_1$, $\hat{A}_1 \in [0,1]^4$, $\mathrm{reason}_1$
    \State $A_1 \gets \text{ratio2xyxy}(\hat{A}_1; W, H)$
    \State $A_1 \gets \textsc{AdjustRoI}(A_1, B^\star)$
    \State Query $\mathcal{G}$ with $(I, q, a, B^\star)$ for a one-sentence final justification
    \State \Return $\{(\mathrm{desc}_1, A_1, \mathrm{reason}_1)\}$ and final justification
\EndIf
\State

\State $t \gets 1$
\State \textbf{(Round 1)} Prompt $\mathcal{G}$ with $(I, q, a)$ to obtain $\mathrm{desc}_1$, $\hat{A}_1 \in [0,1]^4$, $\mathrm{reason}_1$
\State $A_1 \gets \text{ratio2xyxy}(\hat{A}_1; W, H)$
\State $A_1 \gets \textsc{AdjustRoI}(A_1, B^\star)$ \Comment{ensure GT coverage and in-bounds}
\While{$t < R_{\max}$ \textbf{and} $\mathrm{Region}(A_t) > N \cdot \mathrm{Region}(B^\star)$}
    \State $I_{t+1} \gets \text{crop}(I, A_t)$
    \State Prompt $\mathcal{G}$ with $(I_{t+1}, q, a)$ to obtain $\mathrm{desc}_{t+1}$, $\hat{A}_{t+1}$, $\mathrm{reason}_{t+1}$
    \State Map local box to global coords: $A_{t+1} \gets \text{local2global}(\hat{A}_{t+1}; A_t)$
    \State $A_{t+1} \gets \textsc{AdjustRoI}(A_{t+1}, B^\star)$
    \State $t \gets t + 1$
\EndWhile
\State $T \gets t$; $A_{\mathrm{final}} \gets A_T$; $I_{\mathrm{final}} \gets \text{crop}(I, A_{\mathrm{final}})$
\State Query $\mathcal{G}$ once more with $(I_{\mathrm{final}}, q, a, B^\star)$ for a one-sentence final justification
\State Optionally run a consistency check to correct out-of-bound or implausible boxes across $\{A_t\}_{t=1}^{T}$
\State \Return $\{(\mathrm{desc}_t, A_t, \mathrm{reason}_t)\}_{t=1}^{T}$ and final justification

\Function{AdjustRoI}{$A, B^\star$}
  \State $A \gets A \cup B^\star$ \Comment{expand to include the GT box}
  \State $A \gets \text{clipToImage}(A)$
  \State \Return $A$
\EndFunction
\end{algorithmic}
\end{algorithm}

\begin{algorithm}[t]
\caption{\datasetnamemax{}: Depth-aware Spatial CoT Generation}
\label{alg:visreason-pro}
\small
\resizebox{0.92\linewidth}{!}{%
\begin{minipage}{1.0\linewidth}
\begin{algorithmic}[1]

\Require Image $I$; optional seed QA $(q_0, a_0, B^\star_0)$; max rounds $R_{\max}{=}4$; area ratio threshold $N{=}2$; large-object threshold $\tau_{\text{large}}$; QA generator $\mathcal{G}_{\text{Q}}$; CoT generator $\mathcal{G}_{\text{CoT}}$
\Ensure 3D-aware QA $(q, a, B^\star)$ and multi-round chain $\{(\mathrm{desc}_t, \mathrm{AoI}_t, \mathrm{reason}_t)\}_{t=1}^{T}$

\State \textbf{(Pseudo-3D annotation)}
\State $D \gets \text{MonocularDepth}(I)$ \Comment{dense normalized depth map}
\State $S \gets \text{Seg}(I)$ \Comment{instance-level semantic segmentation}
\State Build object list $\mathcal{O} = \{(c_i, [x_1, y_1, x_2, y_2]_i, d_i)\}_{i=1}^{K}$ from $(S, D)$
\State \hspace{1em}\Comment{bounding box, category, and ordinal depth $d_i$ for each object}

\State
\If{no seed QA is provided}
    \State $(q, a, B^\star) \gets \mathcal{G}_{\text{Q}}(I, \mathcal{O})$
    \State \Comment{generate 3D-aware QA with 2D (\emph{left of}, \emph{above}) + depth (\emph{in front of}, \emph{behind}) relations}
\Else
    \State $(q, a, B^\star) \gets (q_0, a_0, B^\star_0)$
\EndIf

\State
\State $(W, H) \gets \text{size}(I)$
\State Compute image area $A_{\text{img}} \gets W \cdot H$ and GT area $A_{\text{GT}} \gets \mathrm{Region}(B^\star)$
\If{$A_{\text{GT}} / A_{\text{img}} \ge \tau_{\text{large}}$}
    \State Prompt $\mathcal{G}_{\text{CoT}}$ with $(I, q, a, \mathcal{O})$ to obtain $\mathrm{desc}_1$, $\hat{A}_1$, $\mathrm{reason}_1$
    \State $A_1 \gets \text{ratio2xyxy}(\hat{A}_1; W, H)$
    \State $A_1 \gets \textsc{AdjustAoI}(A_1, B^\star)$
    \State Optionally distill a single-round rationale from $(\mathrm{desc}_1, \mathrm{reason}_1)$
    \State \Return $(q, a, B^\star)$ and $\{(\mathrm{desc}_1, A_1, \mathrm{reason}_1)\}$
\EndIf

\State
\State $t \gets 1$
\State \textbf{(Round 1)} Prompt $\mathcal{G}_{\text{CoT}}$ with $(I, q, a, \mathcal{O})$ to obtain $\mathrm{desc}_1$, $\hat{A}_1$, $\mathrm{reason}_1$
\State $A_1 \gets \text{ratio2xyxy}(\hat{A}_1; W, H)$
\State $A_1 \gets \textsc{AdjustAoI}(A_1, B^\star)$

\While{$t < R_{\max}$ \textbf{and} $\mathrm{Region}(A_t) > N \cdot \mathrm{Region}(B^\star)$}
    \State $I_{t+1} \gets \text{crop}(I, A_t)$

    \State Extract local object subset $\mathcal{O}_{t+1} \subseteq \mathcal{O}$ whose boxes intersect $A_t$
    \State Re-normalize depths in $\mathcal{O}_{t+1}$ to local ordinal ranks

    \State Prompt $\mathcal{G}_{\text{CoT}}$ with $(I_{t+1}, q, a, \mathcal{O}_{t+1})$ to obtain
    \State \hspace{2em}$\mathrm{desc}_{t+1}$, $\hat{A}_{t+1}$, $\mathrm{reason}_{t+1}$

    \State Map local box to global coords: $A_{t+1} \gets \text{local2global}(\hat{A}_{t+1}; A_t)$
    \State $A_{t+1} \gets \textsc{AdjustAoI}(A_{t+1}, B^\star)$

    \State $t \gets t + 1$
\EndWhile

\State $T \gets t$

\State Optionally distill a single-round rationale from $\{(\mathrm{desc}_t, A_t, \mathrm{reason}_t)\}_{t=1}^{T}$
\State \Return $(q, a, B^\star)$ and $\{(\mathrm{desc}_t, A_t, \mathrm{reason}_t)\}_{t=1}^{T}$

\Function{AdjustAoI}{$A, B^\star$}
  \State $A \gets A \cup B^\star$ \Comment{enforce GT coverage}
  \State $A \gets \text{clipToImage}(A)$
  \State \Return $A$
\EndFunction

\end{algorithmic}
\end{minipage}
}
\end{algorithm}

\clearpage

\section{Evaluation Protocol on Visual-CoT-Benchmark}
\label{appendix:evaluation}

To support the comparisons in Sec. 5.1 in the main paper, we describe here the exact evaluation protocol used on Visual-CoT-Benchmark.

\noindent\textbf{Multi-round zoom-in evaluation.}
To mirror the training-time multi-round spatial CoT supervision, we evaluate models with an explicit \emph{zoom tool} that the model can call when necessary. The tool takes a ratio-based bounding box $\mathrm{bbox}_2{=}[x_1,y_1,x_2,y_2] \in [0,1]^4$ on the \emph{current} view and returns a cropped image patch. Before cropping, we validate the box (all coordinates in $[0,1]$, $x_1<x_2$, $y_1<y_2$); invalid proposals are discarded and the model continues on the previous view. A valid box is mapped to absolute pixel coordinates on the parent view, clipped to the image boundaries, and used to crop the next view. The crop is optionally upsampled (while preserving aspect ratio) to keep its pixel count within the same budget as the global view, and this new patch becomes the input image for the next round.

The dialog protocol is as follows. In the first round, the user message consists of the resized global image and a textual query:
we instruct the model to (i) think before answering, and (ii) either answer directly or request a zoom if the current view is insufficient (small text, heavy clutter, multiple candidates, etc.). If the model decides to zoom, we apply the crop described above, attach the cropped image as the new visual input, and append a short textual instruction indicating that this is a zoomed-in view; the model is again asked to think and decide whether another zoom is needed. This process repeats for at most $R_{\max}{=}5$ rounds per example. As soon as the model produces a final answer (without requesting further zooms), the interaction stops and that answer is recorded. Empirically, most examples terminate within 2–3 rounds.

\noindent\textbf{Answer extraction and dataset handling.}
For each completed interaction, we parse the model output to extract the final natural language answer string. The prompt format explicitly asks the model to wrap its prediction into a dedicated answer segment, but our extractor is robust: if the expected markup is missing, we fall back to the raw text after minimal trimming. For each constituent dataset in Visual-CoT-Benchmark (\textit{e.g.}, CUB, TextVQA, DocVQA, GQA and others), we follow the official validation splits and JSON annotations provided by the benchmark authors. Each entry specifies the image path, the question, and the ground-truth answer (or option letter for multiple-choice questions). For datasets where the original evaluation protocol includes answer normalization (\textit{e.g.}, case folding, punctuation stripping, synonym mapping), we apply the same normalization as the official scripts; we then compute accuracy as the fraction of examples where the normalized prediction matches the normalized ground truth.

\noindent\textbf{Aggregating metrics.}
Per-dataset accuracies are computed independently and then macro-averaged across all Visual-CoT-Benchmark subsets to obtain the “Avg.” scores reported in Tab. 2 in the main paper. All baseline results are either taken from the original Visual-CoT-Benchmark paper or reproduced using the same protocol, ensuring that our \datasetname{}–7B and \datasetnamemax{}–7B numbers are directly comparable to prior MLLMs under an identical evaluation setup.

\section{Additional Evaluation Results}
\label{appendix:additional results}

\subsection{Full Direct-QA Format Analysis}
\label{appendix:direct_qa_format_analysis}

\cref{tab:direct_qa_full_results} reports the full per-dataset comparison between two inference formats for the same \datasetnamemax{}--7B checkpoint: the intended multi-round visual-CoT format and a Direct-QA format following the vanilla Qwen-style short-answer protocol. Direct-QA largely preserves the base model's short-answer behavior on text- and document-centric tasks, whereas the multi-round visual-CoT format provides stronger gains on tasks that require localized evidence, fine-grained recognition, or spatial disambiguation. This suggests that explicit zoom-and-verify reasoning is most useful when the answer depends on spatially localized visual evidence, but may introduce unnecessary overhead for tasks that can already be answered reliably from the global image.

\begin{table}[t]
\centering
\small
\caption{\textbf{Full Direct-QA format analysis.}
We evaluate the same \datasetnamemax{}--7B checkpoint under two inference formats: the intended multi-round visual-CoT format and a Direct-QA format following the vanilla Qwen-style short-answer protocol. Direct-QA preserves the base model's native short-answer behavior on text/document tasks, while the multi-round visual-CoT format yields larger gains on localized fine-grained and spatial tasks. Scores are computed using the same GPT-4o-mini evaluator as in the main paper.}
\label{tab:direct_qa_full_results}
\resizebox{\linewidth}{!}{%
\begin{tabular}{lrrrrrr}
\toprule
Dataset & $n$ & Qwen & \datasetnamemax{} CoT & \datasetnamemax{} Direct & $\Delta_{\mathrm{Direct}}$ & $\Delta_{\mathrm{CoT}}$ \\
\midrule
DocVQA & 846 & 0.964 & 0.928 & 0.959 & -0.005 & -0.036 \\
DUDE & 559 & 0.817 & 0.786 & 0.813 & -0.004 & -0.031 \\
SROIE & 685 & 0.987 & 0.966 & 0.986 & -0.001 & -0.021 \\
TextCaps & 846 & 0.871 & 0.847 & 0.862 & -0.009 & -0.024 \\
TextVQA & 526 & 0.952 & 0.922 & 0.952 & -0.000 & -0.030 \\
CUB & 491 & 0.681 & 0.831 & 0.709 & +0.028 & +0.150 \\
Flickr30k & 1455 & 0.772 & 0.777 & 0.764 & -0.008 & +0.005 \\
Visual7W & 994 & 0.690 & 0.698 & 0.693 & +0.003 & +0.008 \\
GQA & 978 & 0.651 & 0.670 & 0.648 & -0.003 & +0.019 \\
Open Images & 944 & 0.498 & 0.805 & 0.499 & +0.001 & +0.307 \\
VSR & 404 & 0.705 & 0.654 & 0.716 & +0.011 & -0.051 \\
\midrule
Weighted avg. & 8728 & 0.770 & 0.802 & 0.769 & -0.001 & +0.032 \\
\bottomrule
\end{tabular}%
}
\end{table}

\subsection{Second-backbone Validation}
\label{appendix:second_backbone_validation}

\cref{tab:internvl_second_backbone_results} reports the full per-dataset results for the InternVL-2.5-8B transfer experiment summarized in the main paper. After one epoch of fine-tuning on \datasetname{}, the model shows a mixed but informative pattern: performance improves substantially on CUB and Open Images, but decreases on several text/document and spatial-relation subsets. The overall weighted average remains nearly unchanged. We therefore interpret this experiment as evidence that the proposed supervision can transfer to a different backbone, while also indicating that the multi-round format and one-epoch training setting may require backbone-specific calibration.

\begin{table}[t]
\centering
\small
\caption{\textbf{Second-backbone validation on InternVL-2.5-8B.}
We fine-tune InternVL-2.5-8B on \datasetname{} for one epoch and evaluate it on the same Visual-CoT test sets used in the main paper. Scores are GPT-4o-mini judged answer scores in $[0,1]$. Domain and overall averages are weighted by the number of test examples in each dataset.}
\label{tab:internvl_second_backbone_results}
\resizebox{\linewidth}{!}{%
\begin{tabular}{llrrrr}
\toprule
Domain & Dataset & $n$ & InternVL-2.5-8B & + \datasetname{} & $\Delta$ \\
\midrule
\multirow{5}{*}{Text/Doc}
& DocVQA & 846 & 0.846 & 0.797 & -0.049 \\
& TextCaps & 846 & 0.829 & 0.826 & -0.003 \\
& TextVQA & 526 & 0.907 & 0.897 & -0.011 \\
& DUDE & 559 & 0.716 & 0.697 & -0.019 \\
& SROIE & 685 & 0.907 & 0.859 & -0.048 \\
\cmidrule(lr){2-6}
& \textbf{Weighted avg.} & 3462 & \textbf{0.842} & \textbf{0.815} & \textbf{-0.027} \\
\midrule
Fine-grained
& CUB & 491 & 0.747 & 0.823 & +0.076 \\
\cmidrule(lr){2-6}
& \textbf{Weighted avg.} & 491 & \textbf{0.747} & \textbf{0.823} & \textbf{+0.076} \\
\midrule
\multirow{2}{*}{General VQA}
& Flickr30k & 1455 & 0.713 & 0.726 & +0.013 \\
& Visual7W & 994 & 0.681 & 0.622 & -0.059 \\
\cmidrule(lr){2-6}
& \textbf{Weighted avg.} & 2449 & \textbf{0.700} & \textbf{0.683} & \textbf{-0.017} \\
\midrule
\multirow{3}{*}{Spatial Relation}
& GQA & 978 & 0.689 & 0.549 & -0.140 \\
& Open Images & 944 & 0.502 & 0.803 & +0.301 \\
& VSR & 404 & 0.737 & 0.642 & -0.095 \\
\cmidrule(lr){2-6}
& \textbf{Weighted avg.} & 2326 & \textbf{0.621} & \textbf{0.669} & \textbf{+0.047} \\
\midrule
\multicolumn{2}{l}{\textbf{Overall weighted avg.}}
& 8728 & \textbf{0.738} & \textbf{0.740} & \textbf{+0.001} \\
\bottomrule
\end{tabular}%
}
\end{table}

\subsection{Source-held-out / Source-removal Ablation}
\label{appendix:source_removal_ablation}

\cref{tab:source_removal_ablation_results} reports a conservative source-removal ablation in which CUB and spatial-relation training sources are excluded before one-epoch training. This setting is intentionally stringent. Removing domain-relevant sources substantially weakens fine-grained and spatial performance, especially on CUB, GQA, Open Images, Visual7W, and VSR. We therefore use this result as a diagnostic ablation rather than as evidence of broad source-independent generalization. The result indicates that domain-relevant supervision remains important for these categories, and that the gains reported in the main paper should be interpreted as improvements under matched or related visual reasoning sources rather than as fully source-agnostic transfer.

\begin{table}[t]
\centering
\small
\caption{\textbf{Source-removal ablation.}
We remove CUB and spatial-relation training sources and train \datasetname{} for one epoch. This conservative ablation measures how much fine-grained and spatial performance depends on domain-relevant supervision. The large drops on several held-out categories indicate that source-domain coverage remains important.}
\label{tab:source_removal_ablation_results}
\resizebox{\linewidth}{!}{%
\begin{tabular}{lccccccccccc}
\toprule
Setting & CUB & DocVQA & DUDE & Flickr & GQA & OImg. & SROIE & TCap & TVQA & V7W & VSR \\
\midrule
\datasetname{} w/o CUB+Spatial, 1ep
& 0.602 & 0.773 & 0.600 & 0.610 & 0.340 & 0.250 & 0.707 & 0.621 & 0.739 & 0.393 & 0.335 \\
\bottomrule
\end{tabular}%
}
\end{table}

\subsection{Human Audit of Data Quality}
\label{appendix:human_audit}

To further assess annotation quality, we conduct a larger stratified manual audit over 2{,}200 examples. We sample 200 examples from each source sub-dataset and aggregate the results into four domains: Text/Doc, Fine-grained, General VQA, and Spatial Relation. The audit evaluates five aspects: answer consistency, target containment, RoI tightness, rationale necessity, and rationale faithfulness. We report domain-level and overall pass rates with Wilson 95\% confidence intervals in \cref{tab:human_audit_quality}.

\begin{table}[t]
\centering
\small
\caption{\textbf{Human audit of annotation quality.}
We manually audit 2{,}200 examples by sampling 200 examples from each source sub-dataset and aggregating them into four domains. Each cell reports the pass rate with Wilson 95\% confidence interval. The audit covers answer consistency, whether the RoI contains the answer-relevant target, whether the RoI is sufficiently tight, whether the rationale is necessary for answering the question, and whether the rationale is faithful to the visual evidence.}
\label{tab:human_audit_quality}
\resizebox{\linewidth}{!}{%
\begin{tabular}{lcccccc}
\toprule
Domain & $n$ & Ans. & RoI contain & RoI tight & Rat. nec. & Rat. faith. \\
\midrule
Text/Doc
& 1000
& 99.1 [98.3, 99.5]
& 96.6 [95.3, 97.6]
& 90.3 [88.3, 92.0]
& 70.9 [68.0, 73.6]
& 71.9 [69.0, 74.6] \\
Fine-grained
& 200
& 99.0 [96.4, 99.7]
& 100.0 [98.1, 100.0]
& 100.0 [98.1, 100.0]
& 100.0 [98.1, 100.0]
& 99.0 [96.4, 99.7] \\
General VQA
& 400
& 99.5 [98.2, 99.9]
& 100.0 [99.0, 100.0]
& 99.2 [97.8, 99.7]
& 100.0 [99.0, 100.0]
& 99.5 [98.2, 99.9] \\
Spatial Rel.
& 600
& 99.0 [97.8, 99.5]
& 100.0 [99.4, 100.0]
& 98.3 [97.0, 99.1]
& 99.0 [97.8, 99.5]
& 98.0 [96.5, 98.9] \\
\midrule
Overall
& 2200
& 99.1 [98.7, 99.4]
& 98.5 [97.8, 98.9]
& 95.0 [94.0, 95.8]
& 86.5 [85.0, 87.9]
& 86.5 [85.0, 87.9] \\
\bottomrule
\end{tabular}%
}
\end{table}

The audit shows strong answer and localization quality overall: answer consistency reaches 99.1

\subsection{Human Calibration of the GPT-based Evaluator}
\label{appendix:gpt_judge_human_calibration}

We use GPT-4o-mini as the automatic answer-similarity judge for large-scale evaluation. The judge receives the question, the reference answer, and the model prediction, and returns a scalar similarity score in $[0,1]$. The grading prompt is provided in \cref{fig:prompt-gpt-grader}. During evaluation, we run the judge five times per example and average the parsed scalar scores. The parser extracts the numeric value following the required \texttt{score:} field and discards malformed outputs. To verify that this automatic evaluator is aligned with the intended human rubric, we compare GPT-4o-mini scores against human ratings and report the calibration statistics in \cref{tab:gpt_judge_human_alignment}.

\begin{table}[t]
\centering
\small
\caption{\textbf{Human calibration of the GPT-based evaluator.}
We validate GPT-4o-mini against human ratings under the same answer-similarity rubric. The judge is repeated five times per example and the final automatic score is the average parsed scalar score. We report the mean score with 95\% confidence interval, Pearson correlation, Spearman correlation, quadratic weighted kappa (QWK), exact agreement, and agreement within 0.5 points. Exact and within-0.5 agreement are reported as percentages.}
\label{tab:gpt_judge_human_alignment}
\resizebox{\linewidth}{!}{%
\begin{tabular}{lccccccc}
\toprule
Judge & Rep. & Score & Pearson $r$ & Spearman $\rho$ & QWK & Exact (\%) & Within 0.5 (\%) \\
\midrule
Human
& --
& 0.790 [0.759, 0.820]
& --
& --
& --
& --
& -- \\
GPT-4o-mini
& 5
& 0.802 [0.771, 0.832]
& 0.758 [0.696, 0.818]
& 0.894 [0.866, 0.919]
& 0.819 [0.779, 0.868]
& 73.3 [69.4, 76.8]
& 92.5 [90.0, 94.5] \\
\bottomrule
\end{tabular}%
}
\end{table}

The automatic judge closely matches the human score distribution and shows strong alignment with the human rubric, with Pearson correlation $r=0.758$, Spearman correlation $\rho=0.894$, and QWK of 0.819. Exact agreement reaches 73.3\%, and agreement within 0.5 points reaches 92.5\%. These results do not replace human evaluation, but they support GPT-4o-mini as a calibrated and reproducible proxy for large-scale model comparison under our rubric.

\section{Evaluation Protocol on the \datasetnamemax{} Benchmark}
\label{appendix:visreason-pro-eval}

\subsection{Region-of-Interest Localization Evaluation}
\label{appendix:roi-eval}

For the RoI localization metrics in Tab. 3 in the main paper, we evaluate all models under a unified multi-round zoom–answer interface that mimics the \datasetnamemax{} data-generation pipeline.

\noindent\textbf{Multi-round interaction interface.}
Given an image $I$, a question $q$, and its ground-truth target box $B^\star$, the model is queried in a chat-style interface with tool support. The system prompt exposes a single image \emph{zoom} tool whose argument is a ratio-based bounding box $\mathrm{bbox}_2{=}[x_1,y_1,x_2,y_2]\in[0,1]^4$ on the \emph{current view}. At each round, the model must always produce a private reasoning segment enclosed in \texttt{<think>...</think>}. It can then either:
(i) call the zoom tool once via a \texttt{<tool\_call>} tag, or
(ii) output the final answer via an \texttt{<answer>} tag.
The prompt strictly forbids mixing \texttt{<tool\_call>} and \texttt{<answer>} in the same round, ensuring a clean separation between “zoom” and “answer” steps. We cap the total number of rounds by a small constant $R_{\max}$ (typically $R_{\max}{=}6$).

\noindent\textbf{Zoom execution and box propagation.}
Whenever the model emits a \texttt{<tool\_call>}, we parse the JSON arguments, validate the predicted ratio box (four numeric coordinates, all in $[0,1]$, with $x_1\!<\!x_2$, $y_1\!<\!y_2$), and map it to absolute pixel coordinates on the current view. The resulting box is clipped to the image boundaries and used to crop the next view; the crop is resized to satisfy a fixed pixel budget while preserving aspect ratio, and encoded back into the next-round prompt as an image. This process is repeated until either the model chooses to answer (emits \texttt{<answer>}) or the round budget $R_{\max}$ is reached. Any malformed or out-of-range box proposals are discarded and the model continues with the previous view, which prevents pathological tool calls from corrupting the evaluation.

\noindent\textbf{Predicted RoI extraction.}
For models trained with multi-round spatial supervision (our \datasetnamemax{}–7B and \datasetname{}–7B), the final predicted RoI is taken to be the last valid zoom box produced in the interaction, \textit{i.e.}, the box that defines the final view when the answer is given. For single-step baselines that do not explicitly emit zoom calls, the prompt instructs them to directly produce a single ratio-based box in their first (and only) reasoning step, which is then treated as their predicted RoI. In both cases, the predicted box is represented in the global image coordinate system to allow direct comparison with $B^\star$.

\noindent\textbf{Localization metrics.}
Given the predicted box $\hat{B}$ and the ground-truth box $B^\star$, we compute the intersection-over-union score
\[
  \mathrm{IoU}(\hat{B},B^\star)
  = \frac{\mathrm{Area}(\hat{B}\cap B^\star)}{\mathrm{Area}(\hat{B}\cup B^\star)}.
\]
RoI accuracy is then reported as the fraction of examples whose IoU exceeds standard thresholds, \textit{i.e.}, $\mathrm{IoU@0.5}$ and $\mathrm{IoU@0.75}$. Our models use the refined last-round box $\hat{B}_T$ after multi-round zoom-and-verify, whereas baselines rely on their single-step prediction, making improvements attributable to our iterative cropping strategy rather than differences in IoU computation.

\subsection{3D Grounding Evaluation}
\label{appendix:3d-grounding-eval}

For the 3D grounding metrics in Tab. 4 in the main paper, we evaluate whether models can \emph{explicitly} annotate key objects with consistent 2D boxes and ordinal depth in their reasoning traces.

\noindent\textbf{Reasoning-format constraints.}
Each evaluation example consists of an image $I$, a question $q$, and pseudo-3D annotations from \datasetnamemax{}, which include a set of key objects with ground-truth ratio boxes and normalized depths. At inference time, the system prompt requires every model to write, \emph{inside} its \texttt{<think>} segment, a list of spatial annotations of the form
\[
\text{name}: ([x_1, y_1, x_2, y_2], \text{depth}),
\]
for all salient or answer-relevant objects. The box coordinates must be ratio-based and lie in $[0,1]$, while the depth is also constrained to $[0,1]$ (interpreted as a discrete ordinal depth from the current viewpoint). The model is free to choose object names, but is explicitly encouraged to annotate the objects it relies on to answer the question. No additional crops are allowed in this protocol; we focus purely on the quality of grounding expressed in the reasoning itself.

\noindent\textbf{Parsing predicted groundings.}
From the generated \texttt{<think>} segment, we parse all entries matching the pattern
\[
\text{name}: ([x_1, y_1, x_2, y_2], d),
\]
using a robust regular expression that tolerates different separators, whitespace, and minor formatting variations. The parser supports: (i) boxes given directly as ratios; (ii) boxes given in percentages (automatically divided by $100$); and (iii) boxes given in pixels (optionally normalized using the image width and height, when available). All boxes are clipped to $[0,1]^4$ and corrected to ensure $x_1<x_2$, $y_1<y_2$. Depth values $d$ can be numeric or symbolic; numeric depths are normalized to $[0,1]$, while symbolic depths such as “near”, “mid(dle)”, and “far” are mapped to fixed ordinal levels (\textit{e.g.}, $0.2$, $0.5$, $0.8$). The result is, for each model, a dictionary of predicted objects with fields \texttt{bbox\_ratio} and \texttt{depth01}.

\noindent\textbf{Ground-truth object set.}
On the dataset side, \datasetnamemax{} provides, for each reasoning step, the key objects that should be grounded in order to support the answer, along with their ratio boxes and normalized depths. For the analysis in Tab. 4 in the main paper, we focus on the first reasoning step (which is shared across all models) and extract the ground-truth object set by parsing the reference description and rationale with the same grounding parser, yielding a dictionary of ground-truth objects with \texttt{bbox\_ratio} and \texttt{depth01} fields.

\noindent\textbf{Matching and metrics.}
For each example, we align predicted and ground-truth objects by their textual names (after lowercasing). Let $\mathcal{G}$ be the set of ground-truth object names and $\mathcal{P}$ the set of predicted names; the intersection $\mathcal{M} = \mathcal{G} \cap \mathcal{P}$ defines matched objects, while $\mathcal{G}\setminus\mathcal{P}$ are missed. For every $o\in\mathcal{M}$, we compute:
\begin{itemize}
  \item the IoU between the predicted and ground-truth boxes,
        $\mathrm{IoU}_o = \mathrm{IoU}(\hat{B}_o, B^\star_o)$ in ratio space; and
  \item the absolute depth error,
        $|\hat{d}_o - d^\star_o|$,
        where both depths are in $[0,1]$.
\end{itemize}
At the example level, we record: (i) the number of ground-truth objects $|\mathcal{G}|$; (ii) the number of missed objects $|\mathcal{G}\setminus\mathcal{P}|$; (iii) the mean IoU over matched objects; and (iv) the mean absolute depth error over matched objects. Aggregating over the entire benchmark yields:
\begin{itemize}
  \item the \emph{grounded ratio}, defined as $1 - \frac{\sum \text{missed}}{\sum |\mathcal{G}|}$, \textit{i.e.}, the fraction of ground-truth objects that are explicitly grounded by the model;
  \item the overall mean IoU across all matched objects; and
  \item the overall mean absolute depth error across all matched objects.
\end{itemize}
These metrics jointly capture whether models (i) choose to ground the right objects in their reasoning, (ii) localize them precisely in 2D, and (iii) assign consistent ordinal depths. Since the protocol never prescribes \emph{which} objects must be grounded in advance, high scores indicate that spatial grounding emerges naturally as part of the model’s reasoning process, rather than being enforced by the evaluation interface.

\section{Prompts Design}
\label{appendix:prompts}

We provide the prompts used in dataset generation and GPT-evaluation in \cref{fig:prompt-system-2d}, \cref{fig:prompt-first-2d}, \cref{fig:prompt-later-2d}, \cref{fig:prompt-fullimage-2d}, \cref{fig:prompt-3dqa}, \cref{fig:prompt-3d-round}, \cref{fig:prompt-3d-distill}, \cref{fig:prompt-3d-grounding}, \cref{fig:prompt-gpt-grader}.

\section{More Data Examples}
\label{appendix:more data}
\begin{figure}[h!]
\centering
\includegraphics[width=0.65\linewidth]{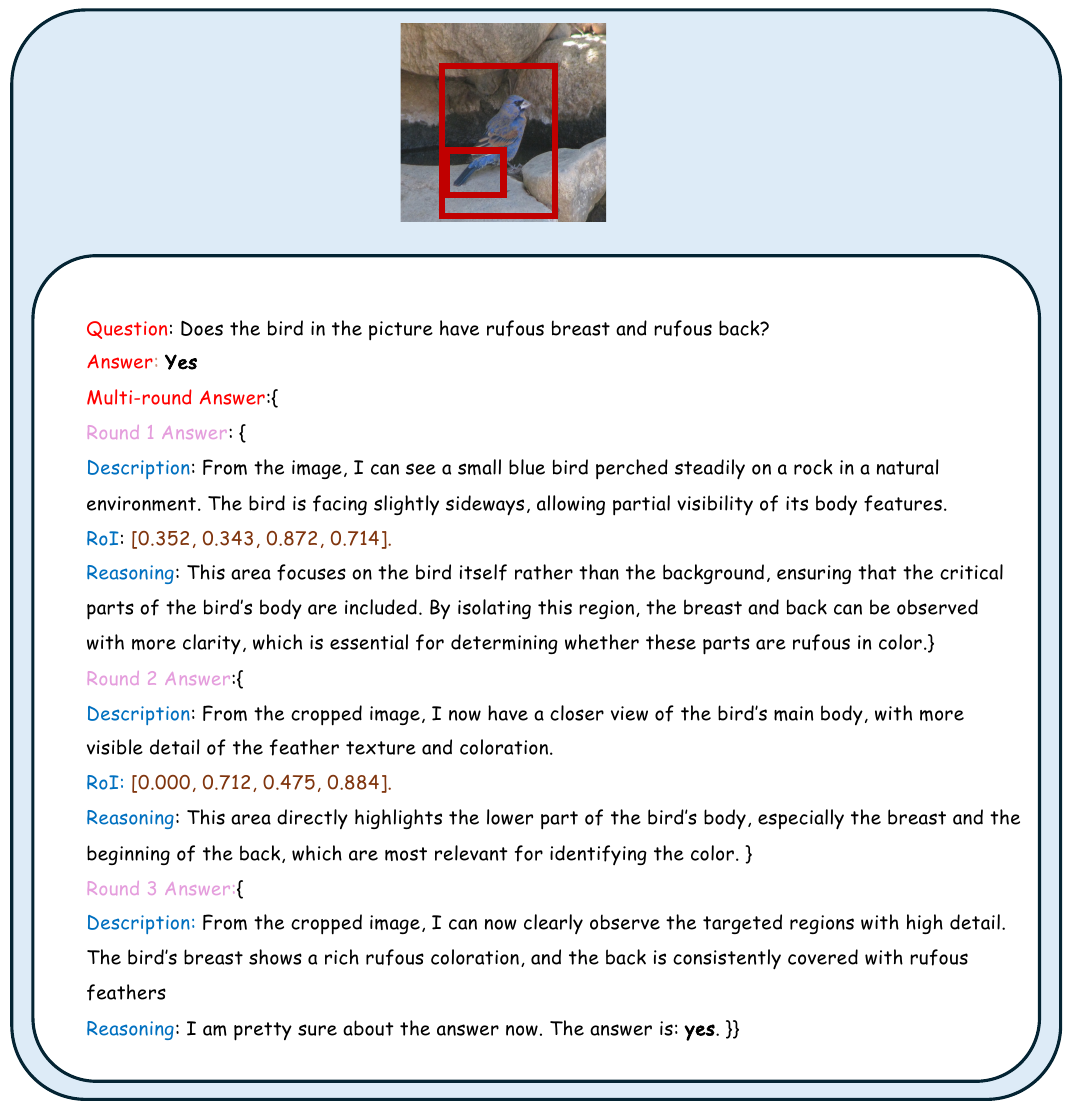}
\caption{More data examples in \modelname{}.
}
\label{fig:data_example_1}
\end{figure}

\begin{figure}[h!]
\centering
\includegraphics[width=0.65\linewidth]{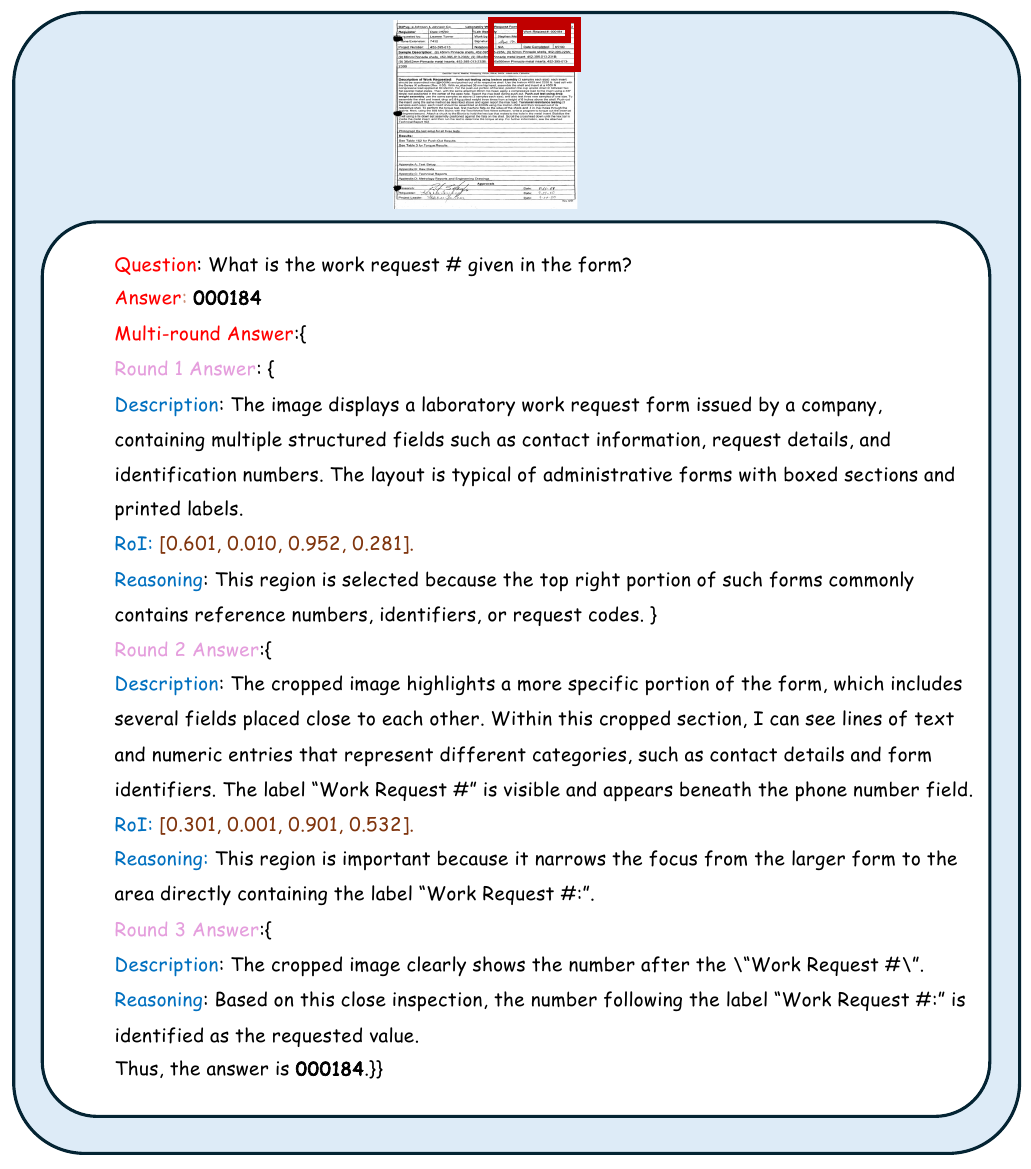}
\caption{More data examples in \modelname{}.
}
\label{fig:data_example_2}
\end{figure}

\begin{figure}[h!]
\centering
\includegraphics[width=0.65\linewidth]{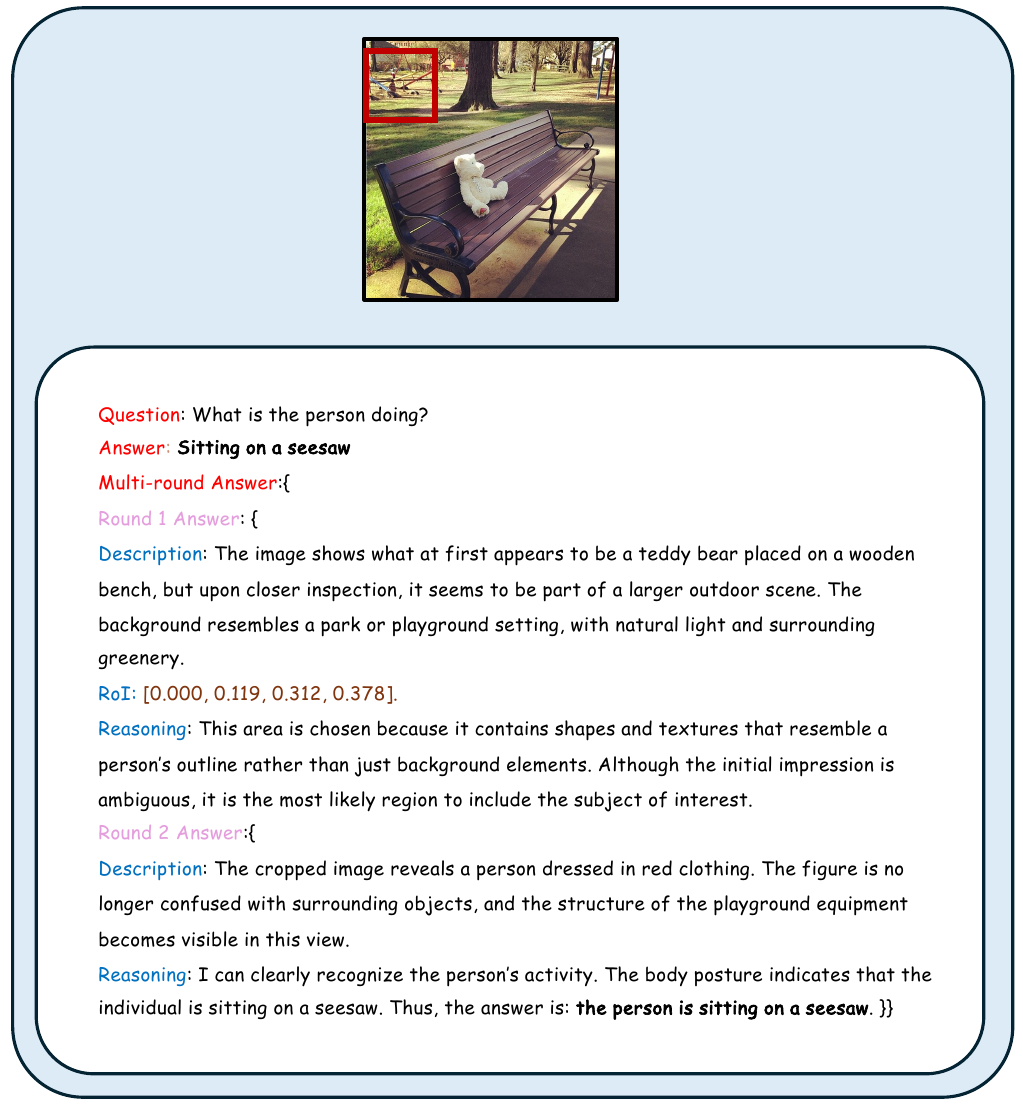}
\caption{More data examples in \modelname{}.
}
\label{fig:data_example_3}
\end{figure}

\begin{figure}[h!]
\centering
\includegraphics[width=0.65\linewidth]{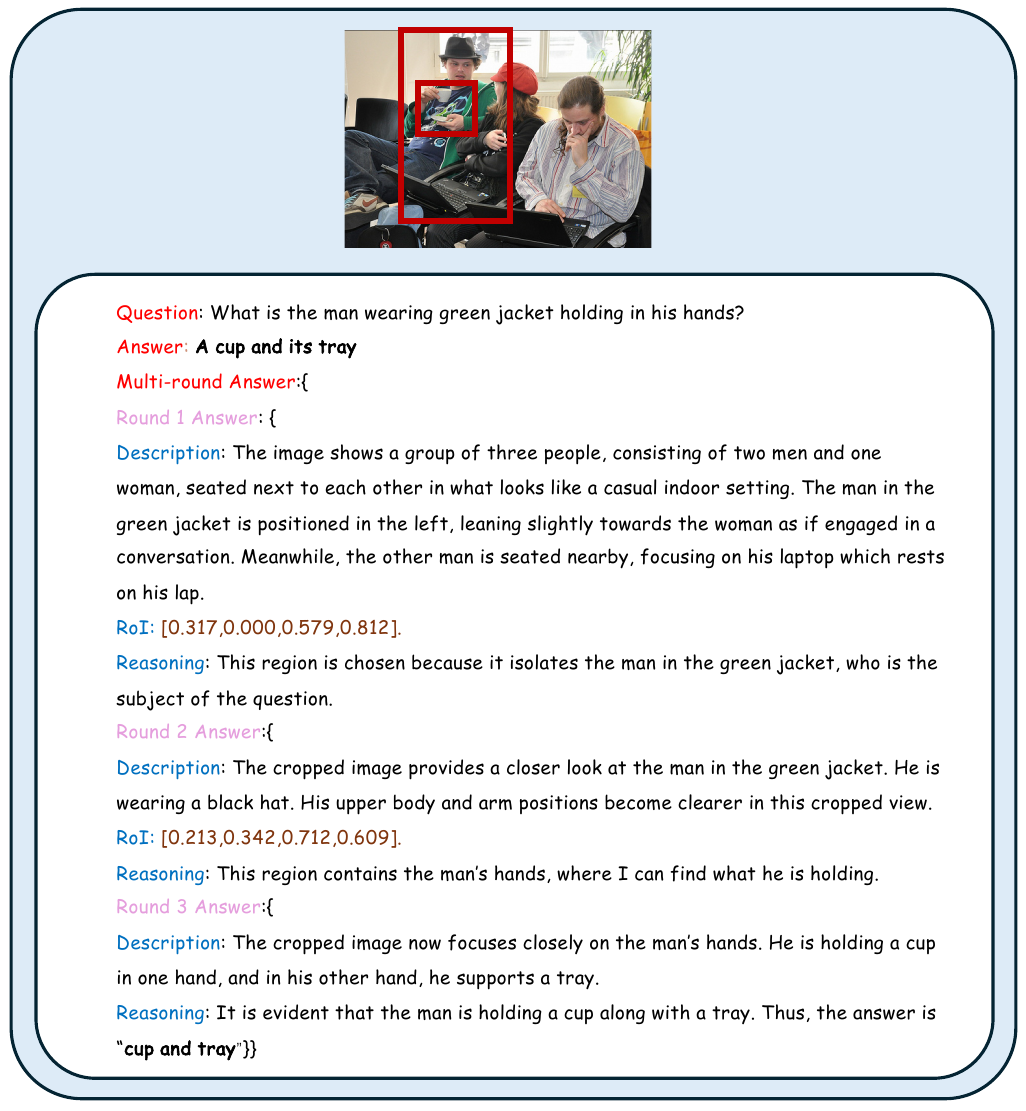}
\caption{More data examples in \modelname{}.
}
\label{fig:data_example_4}
\end{figure}

We show more examples in our dataset in \cref{fig:data_example_1}, \cref{fig:data_example_2}, \cref{fig:data_example_3}, and \cref{fig:data_example_4}.

\section{More Inference Examples}
\label{appendix:more inference exmaples}

\begin{figure}[h!]
\centering
\includegraphics[width=0.75\linewidth]{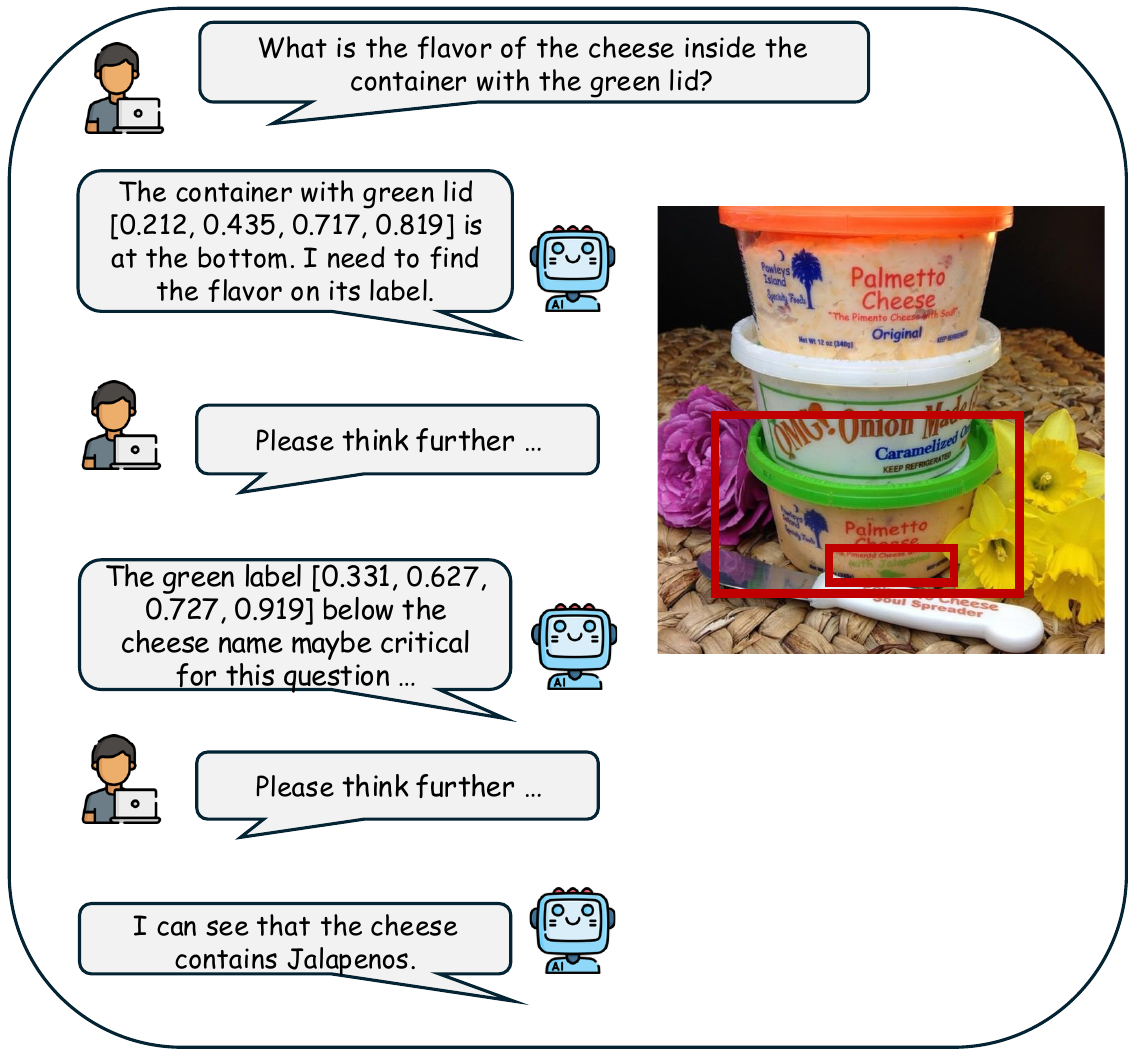}
\caption{More inference examples in \modelname{}.
}
\label{fig:inference_demo_1}
\end{figure}

\begin{figure}[h!]
\centering
\includegraphics[width=0.75\linewidth]{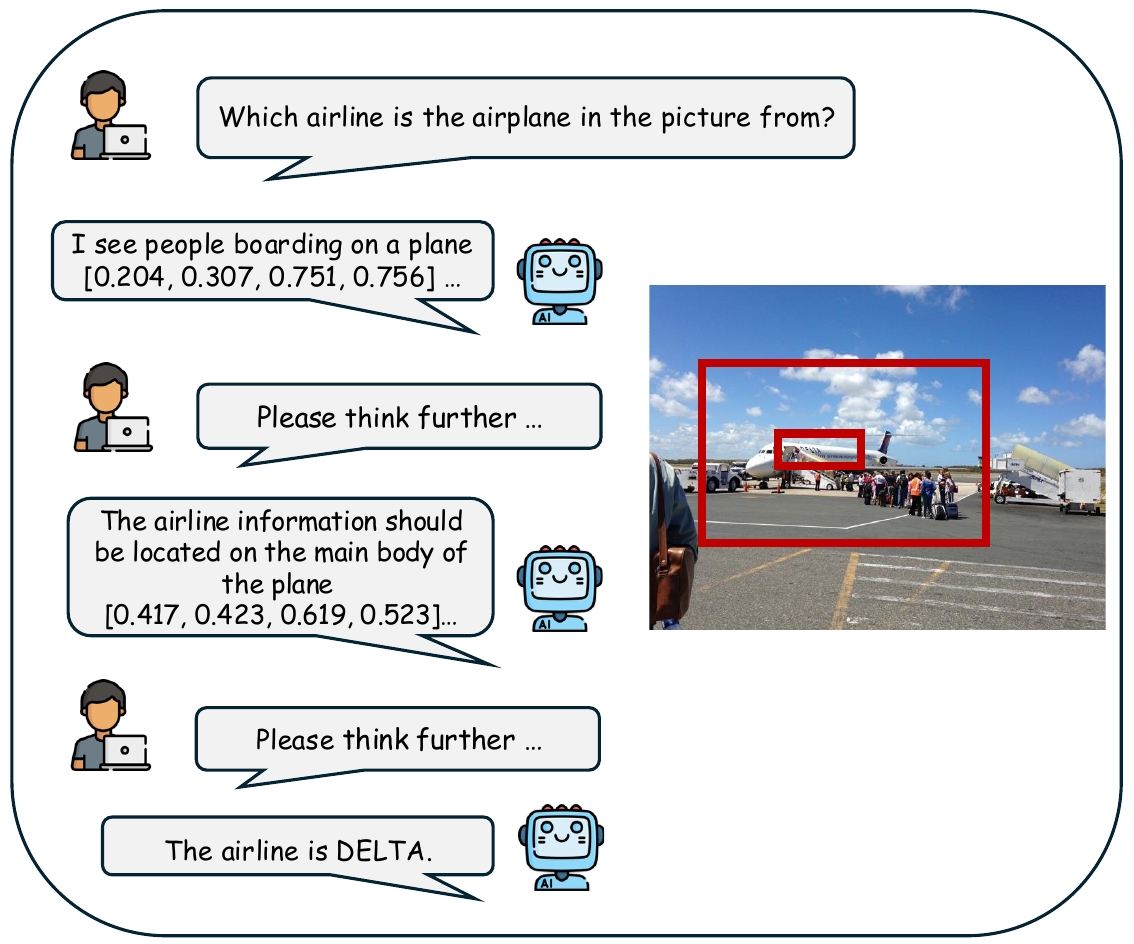}
\caption{More inference examples in \modelname{}.
}
\label{fig:inference_demo_2}
\end{figure}

\begin{figure}[h!]
\centering
\includegraphics[width=0.75\linewidth]{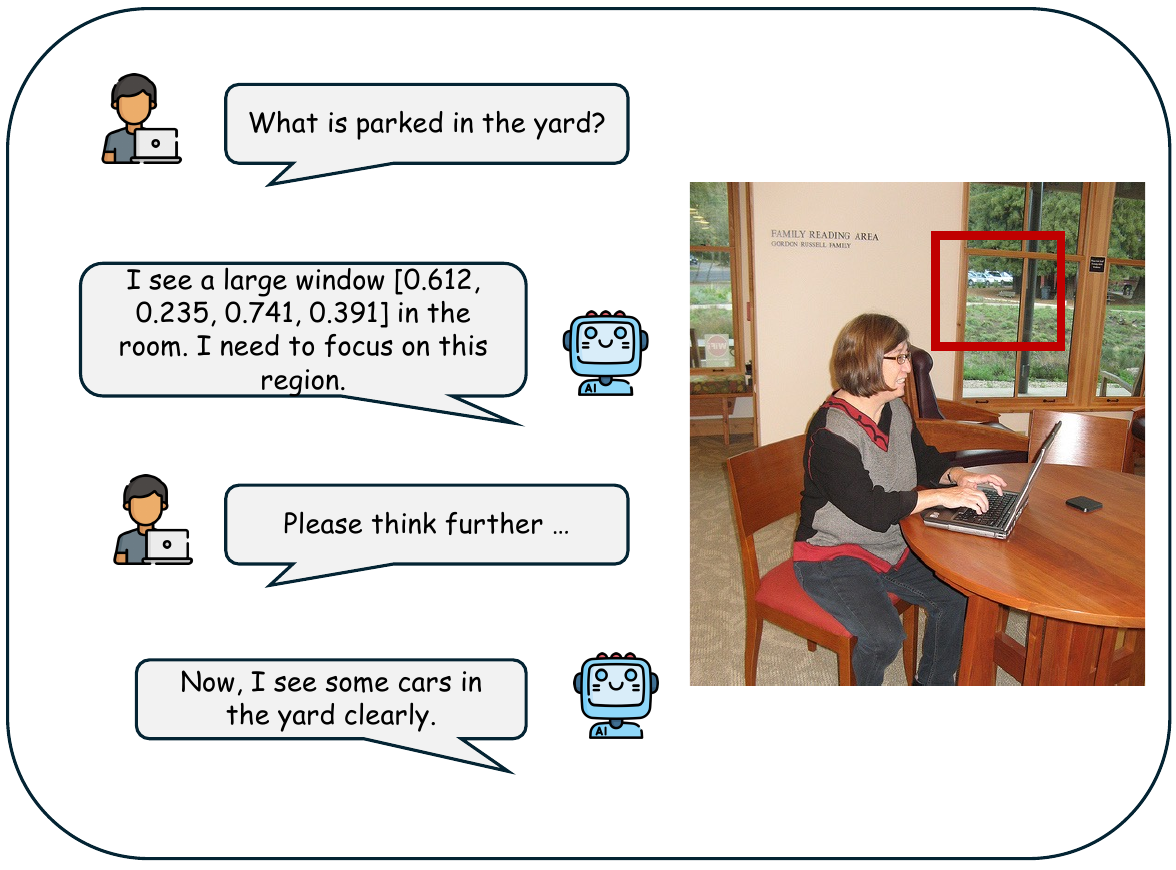}
\caption{More inference examples in \modelname{}.
}
\label{fig:inference_demo_3}
\end{figure}

We provide more inference examples in \cref{fig:inference_demo_1}, \cref{fig:inference_demo_2}, and \cref{fig:inference_demo_3}.

\clearpage



\begin{figure}[t]
\centering
\begin{minipage}{0.97\linewidth}
\lstset{
  breaklines=true,
  breakatwhitespace=true,
  basicstyle=\ttfamily\scriptsize, 
  columns=fullflexible,
  frame=single,
}
\begin{lstlisting}
You are a helpful assistant.

# Tools
You may call one or more functions to assist with the user query.
You are provided with function signatures within <tools></tools> XML tags:
<tools>
{"type":"function",
 "function":{
   "name":"image_zoom_in_tool",
   "description":"Zoom in on a specific region of an image by cropping it 
                  based on a RATIO-BASED bounding box (bbox_2d). 
                  IMPORTANT: All coordinates must be ratios between 0 
                  and 1, not absolute pixel values.",
   "parameters":{
      "properties":{
         "bbox_2d":{
             "type":"array",
             "items":{"type":"number"},
             "minItems":4,
             "maxItems":4,
             "description":"The bounding box of the region to zoom in, 
                            as [x1,y1,x2,y2], where (x1,y1) is the 
                            top-left corner and (x2,y2) is the 
                            bottom-right. All values MUST be ratios 
                            between 0 and 1."
         }
      },
      "required":["bbox_2d"],
      "type":"object"
   },
   "args_format":"Format the arguments as a JSON object."
 }}
</tools>

## STRICT FORMATTING
- Always wrap private reasoning in <think>...</think>.
- If zooming: output EXACTLY ONE <tool_call>...</tool_call>.
- If NOT zooming: output BOTH <think>...</think> and <answer>...</answer>.
- Never mix <answer> with <tool_call>.
- Output nothing outside these tags.

## DECISION RULE (when to call the tool)
Call the zoom tool ONLY IF at least one is true:
1) The target text/object is too small or low-resolution.
2) Counting/attribute verification requires isolating a region.
3) Multiple candidate regions exist and must be disambiguated.

Do NOT call the tool when:
- The answer is already visible.
- The question is answerable from global context.
- Additional pixels would not give new information.

## OUTPUT STYLE
- Be concise and answer exactly the question.
- Never mix <answer> with <tool_call>.
\end{lstlisting}
\end{minipage}
\caption{System prompt used for 2D VisReason generation. The model decides whether zoom-in is needed and must follow strict XML-style reasoning and tool usage rules.}
\label{fig:prompt-system-2d}
\end{figure}

\begin{figure}[t]
\centering
\begin{minipage}{0.97\linewidth}
\lstset{
  breaklines=true,
  breakatwhitespace=true,
  basicstyle=\ttfamily\small,
  columns=fullflexible,
  frame=single,
}
\begin{lstlisting}
Answer the question by first thinking.

If a tighter crop is required, call image_zoom_in_tool once.
If the current view is sufficient, produce <think>...</think>
followed immediately by <answer>...</answer>.

Formatting (must follow strictly):
- Always include <think>...</think>.
- If zooming: output exactly one <tool_call>...</tool_call>.
- If NOT zooming: output <think>...</think> + <answer>...</answer>.
- Never mix <answer> with <tool_call>.
- Output nothing outside these tags.
\end{lstlisting}
\end{minipage}
\caption{First-round prompt for VisReason. The model must decide whether the full image is sufficient or whether zooming is required.}
\label{fig:prompt-first-2d}
\end{figure}

\begin{figure}[t]
\centering
\begin{minipage}{0.97\linewidth}
\lstset{
  breaklines=true,
  breakatwhitespace=true,
  basicstyle=\ttfamily\small,
  columns=fullflexible,
  frame=single,
}
\begin{lstlisting}
Think first. If a tighter crop is required, call image_zoom_in_tool
ON THE CURRENT VIEW (result of previous crop).

If the area is already sufficient, produce reasoning + final answer.

Formatting:
- Always include <think>...</think>.
- If zooming: EXACTLY ONE <tool_call>...</tool_call>.
- If NOT zooming: <think>...</think> + <answer>...</answer>.
- Never mix <answer> with <tool_call>.
\end{lstlisting}
\end{minipage}
\caption{Later-round prompt for VisReason. The model continues zoom-in refinement only when necessary.}
\label{fig:prompt-later-2d}
\end{figure}

\begin{figure}[t]
\centering
\begin{minipage}{0.97\linewidth}
\lstset{
  breaklines=true,
  basicstyle=\ttfamily\small,
  frame=single,
  columns=fullflexible
}
\begin{lstlisting}
You are given:
- An image showing a complex scene.
- A list of all objects in the image (objects_ratio).
  Each object includes its category, bounding box [x1,y1,x2,y2],
  and its depth value.

Your task is:
1. Provide a natural scene description.
2. Predict an Area of Interest (AoI) formatted as [x1,y1,x2,y2] in [0,1].
   The AoI must:
   - Strictly cover the object(s) relevant to the question;
   - Include necessary context;
   - Avoid unrelated areas.
3. Provide a brief reasoning step explaining why this area is sufficient.

Output format:

Scene Description:
[your description]

Area of Interest:
[x1, y1, x2, y2]

Reasoning:
[your explanation]
\end{lstlisting}
\end{minipage}
\caption{Prompt used for first-round full-image RoI generation in 2D VisReason.}
\label{fig:prompt-fullimage-2d}
\end{figure}

\begin{figure}[t]
\centering
\begin{minipage}{0.97\linewidth}
\lstset{
  breaklines=true,
  basicstyle=\ttfamily\small,
  frame=single,
  columns=fullflexible
}
\begin{lstlisting}
You are given an image with:
- A list of objects {category, bounding box [x1,y1,x2,y2], depth value};
- Monocular depth map + segmentation-derived object ordering.

Your task:
Generate a 3D-aware question-answer pair (q, a, B*) that:
- Combines 2D relations (left of, right of, above, below)
  WITH depth relations (in front of, behind);
- Identifies a single target object and produces its GT box B*;
- Avoids ambiguous or underspecified relations;
- Produces a question solvable from the image alone.
\end{lstlisting}
\end{minipage}
\caption{Prompt for generating 3D-aware QA pairs used in VisReason-Pro construction.}
\label{fig:prompt-3dqa}
\end{figure}

\begin{figure}[t]
\centering
\begin{minipage}{0.97\linewidth}
\lstset{
  breaklines=true,
  basicstyle=\ttfamily\small,
  frame=single,
  columns=fullflexible
}
\begin{lstlisting}
Given: image crop, question q, answer a, and local object list O_t.
Each object includes:
- category,
- bounding box [x1,y1,x2,y2],
- ordinal depth.

Task:
1. Provide a concise scene description based on the CURRENT CROP.
2. Predict an area of interest (AoI) [x1,y1,x2,y2] in [0,1].
3. Provide a brief reasoning step.

Rules:
- AoI must cover the target object and supporting evidence.
- Use depth information to disambiguate occlusions or ordering.
- Use <think> for reasoning and <answer> for explanation.
\end{lstlisting}
\end{minipage}
\caption{3D-aware CoT step prompt used in VisReason-Pro for each zoom-in round.}
\label{fig:prompt-3d-round}
\end{figure}

\begin{figure}[t]
\centering
\begin{minipage}{0.97\linewidth}
\lstset{
  breaklines=true,
  basicstyle=\ttfamily\small,
  frame=single,
  columns=fullflexible
}
\begin{lstlisting}
Given multi-round traces {desc_t, AoI_t, reason_t}, distill them into
a SINGLE compact reasoning step:

- Summarize the multi-round logic into one brief explanation.
- Output ONE final AoI box in ratio coordinates [x1,y1,x2,y2].
- Ensure the box tightly covers the correct target object.
- Avoid mentioning intermediate zooms, crops, or recursion.
\end{lstlisting}
\end{minipage}
\caption{Prompt for single-round distillation in VisReason-Pro. Converts multi-round chains into a compact one-shot explanation.}
\label{fig:prompt-3d-distill}
\end{figure}

\begin{figure}[t]
\centering
\begin{minipage}{0.97\linewidth}
\lstset{
  breaklines=true,
  basicstyle=\ttfamily\small,
  frame=single,
  columns=fullflexible
}
\begin{lstlisting}
You are given a piece of reasoning text and a list of visible objects 
with grounding annotations.

Visible objects (JSON):
{objects_json}

Task:
- Rewrite the reasoning text by inserting grounding right AFTER each 
  mentioned key object.
- Grounding format: object_name ([x1, y1, x2, y2], depth_value).
- If depth_value is null or missing, output only ([x1, y1, x2, y2]).
- DO NOT change wording or add new information. Preserve the original 
  reasoning flow.
- DO NOT add objects that are not mentioned.
- Return only the rewritten text.

Text:
{base_text}
\end{lstlisting}
\end{minipage}
\caption{Prompt for grounding augmentation during \datasetnamemax{} generation. 
The model rewrites the reasoning text by inserting explicit spatial grounding 
\texttt{([x1, y1, x2, y2], depth)} after each referenced object without 
altering the semantic content.}
\label{fig:prompt-3d-grounding}
\end{figure}

\begin{figure}[t]
\centering
\begin{minipage}{0.97\linewidth}
\lstset{
  breaklines=true,
  basicstyle=\ttfamily\small,
  frame=single,
  columns=fullflexible
}
\begin{lstlisting}
You are an automatic grader. Given Question / Standard answer / Model's answer,
output a similarity score in [0,1] and nothing else, formatted exactly as:
score: <score> (up to two decimals).

Rules:
- Normalize case and whitespace before comparing.
- Unify number and time formats when they are semantically equivalent.
- If the model answer is exactly the same as the standard answer
  (or only differs in trivial formatting), output score: 1.00.
- If the two answers are synonyms or one clearly contains the other
  without contradiction, output a score in the range 0.95-1.00.
- If there are minor wording differences but the core meaning is the same,
  use a score around 0.85-0.94.
- If there is only partial overlap in meaning, use a score around 0.75-0.84.
- If the answer is unrelated, contradictory, or clearly wrong,
  use a score close to 0.00.

For numerical answers:
- Use smaller relative error for higher scores.
- If both are intervals, use larger overlap for higher scores.
Grade only the core answer, ignoring polite phrases or extra commentary.
\end{lstlisting}
\end{minipage}
\caption{Prompt used for GPT-based similarity grading. The grader receives the question, standard answer, and model prediction, and returns a single scalar score in \([0,1]\) following the specified rubric.}
\label{fig:prompt-gpt-grader}
\end{figure}

\end{document}